\documentclass[final,5p,times,twocolumn,authoryear]{elsarticle}
\usepackage[linesnumbered,ruled,vlined]{algorithm2e}
\usepackage{amssymb}
\usepackage{graphicx}
\usepackage{url}

\newcommand{\ie}{{\emph{i.e.},} }

\usepackage{subfigure}
\usepackage{amsmath}

\usepackage{amssymb}
\usepackage{xcolor}
\usepackage{tabularx} 
\usepackage{multirow}
\usepackage{pbox}
\usepackage{booktabs}

\usepackage{amssymb}
\usepackage{amsfonts}
\usepackage{bbm}
\usepackage{graphicx}
{\begin{list}               
    {$\bullet$ \hfill}{
        \setlength{\leftmargin}{\parindent}
        \setlength{\parsep}{0.04\baselineskip}
        \setlength{\itemsep}{0.5\parsep}
        \setlength{\labelwidth}{\leftmargin}
        \setlength{\labelsep}{0em}}
    }
{\end{list}}

\providecommand{\cref}[1]{Chapter~\ref{#1}}
\providecommand{\sref}[1]{Section~\ref{#1}}
\providecommand{\fref}[1]{Figure~\ref{#1}}

\renewcommand{\vec}[1]{\ensuremath{\boldsymbol{#1}}}
\providecommand{\mat}[1]{\ensuremath{\boldsymbol{#1}}}

\providecommand{\calP}{\mathcal{P}}

\providecommand{\calS}{\mathcal{S}}

\providecommand{\mI}{\mat{I}}

\providecommand{\mM}{\mat{M}}

\providecommand{\vp}{\vec{p}}

\journal{Computers and Electronics in Agriculture}

\begin{document}

\begin{frontmatter}

\title{Learning Adversarial Augmentation Policies for Robust Garlic Seedling Detection}

\author[1]{Soeun Lee}
\ead{dlths67@cau.ac.kr}

\author[1]{Chanho Kim}
\ead{kch2944512@cau.ac.kr}

\author[2]{Yeji Kang}
\ead{yeji930@korea.kr}

\author[2]{YoungKi Hong}
\ead{sanm70@korea.kr}

\author[1]{Byeongkeun Kang\corref{cor1}}
\ead{byeongkeunkang@cau.ac.kr}
\cortext[cor1]{Corresponding author.}
\affiliation[1]{organization={School of Electrical and Electronics Engineering, Chung-Ang University},
            addressline={84 Heukseok-ro, Dongjak-gu}, 
            city={Seoul},
            postcode={06974}, 
            country={South Korea}}
\affiliation[2]{organization={Department of Agricultural Engineering, National Institute of Agricultural Sciences},
            addressline={310 Nongsaengmyeong-ro, Deokjin-gu, Jeonju-si, Jeonbuk-do}, 
            postcode={54875}, 
            country={South Korea}}

\begin{abstract}
Accurate seedling detection during early growth stages is essential for timely replanting and effective crop management in precision agriculture. However, existing seedling detection studies are mostly evaluated under relatively stable imaging conditions, such as UAV imagery or greenhouse environments, leaving robust detection under severe and spatially heterogeneous illumination in ground-based outdoor monitoring insufficiently explored. In addition, many illumination-robust detection methods rely on additional enhancement or feature-extraction modules, which can increase inference-time overhead and are not specifically designed for seedling detection and downstream missing seedling localization. To address these gaps, we construct a new garlic seedling dataset captured using a ground-based monitoring platform under real outdoor field conditions with highly variable illumination. We further propose an illumination-robust seedling detection framework based on adversarial augmentation policy learning. The proposed method jointly optimizes a stochastic augmentation policy agent and an object detector, enabling the detector to learn robust representations under challenging visual conditions. The policy agent generates input-conditioned augmentation strategies that introduce adversarial perturbations during training. To prevent unrealistic distortions, we incorporate a structural penalty into the reward function to preserve the structural consistency of the images while encouraging challenging augmentations. Extensive experiments demonstrate that the proposed approach achieves an AP$_{50}$ of 91.6\%, improving the baseline by 0.9 percentage points and outperforming the previous best-performing method by 0.2 percentage points. For downstream missing seedling localization, the proposed method achieves 75.0\% precision and a 67.0\% F1-score, improving the baseline by 4.8 and 2.0 percentage points, respectively. Compared with the best prior methods, it further improves missing seedling precision by 4.0 percentage points and F1-score by 2.3 percentage points. These results highlight the effectiveness of the proposed framework for practical ground-based agricultural monitoring under complex outdoor lighting conditions without additional inference-time computational overhead.
\end{abstract}

\end{frontmatter}

\section{Introduction}
Accurate identification of seedlings during the early growth stage plays a vital role in supporting timely replanting operations and improving overall crop productivity. In real agricultural practices, gaps in seedling emergence often arise due to issues such as transplanter malfunction, unsuccessful germination, or unfavorable environmental conditions.

To address this challenge, numerous techniques have been proposed for detecting seedlings across various crop species and cultivation settings~\citep{Gennaro_2020_evaluation, Gao_2022_design, CHEN_2022_assimilation, Yuan_2024_rapidly}. In recent years, data-driven approaches, particularly learning-based models, have attracted increasing attention for enhancing detection performance and robustness under diverse field conditions~\citep{YAN_2023_machine, Cui_2023_real_time, Wu_2025_novel}.

\begin{figure*}[t]
\centering
\begin{minipage}{0.25\linewidth}
\centering
\includegraphics[width=\linewidth]{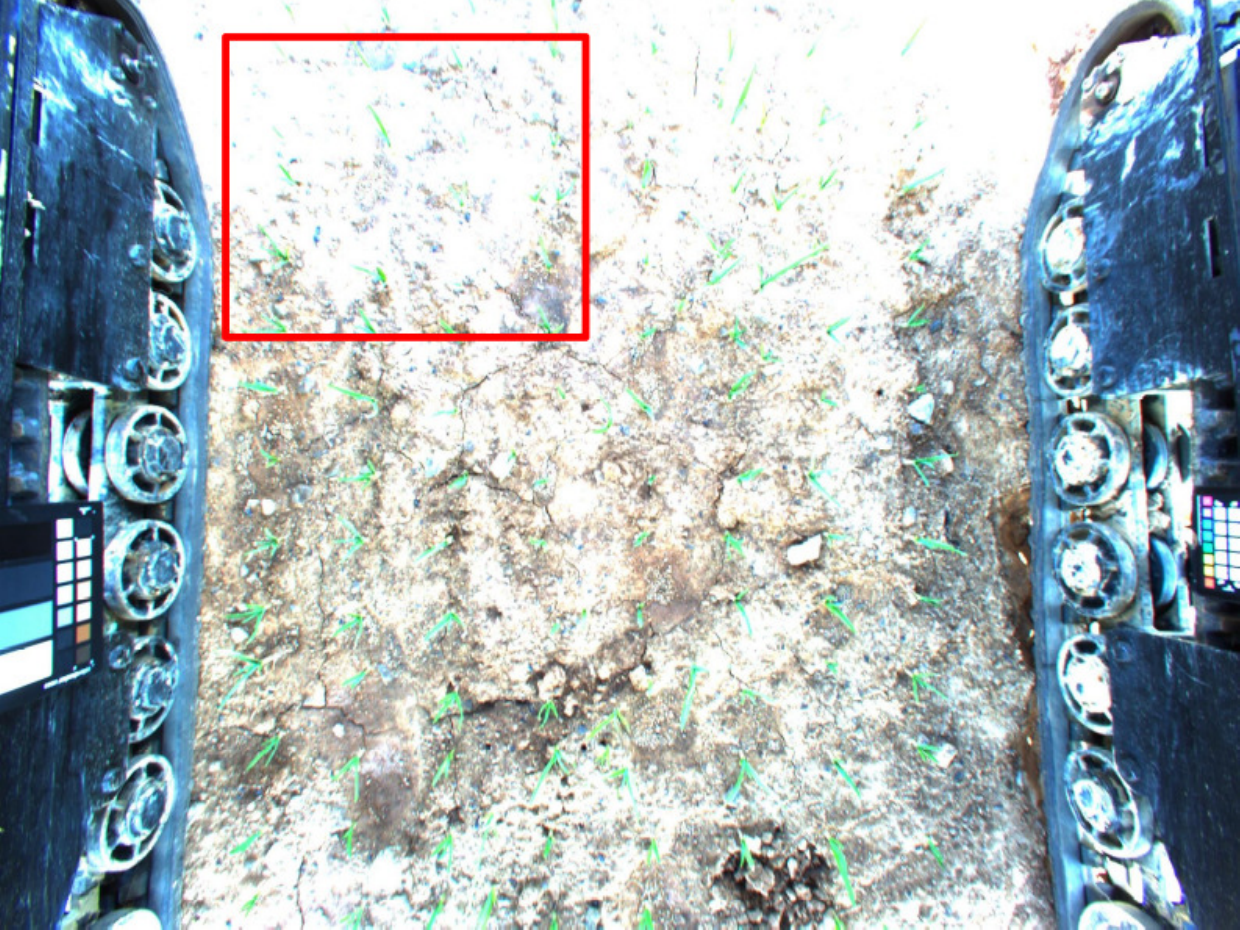}
\end{minipage}
\begin{minipage}{0.25\linewidth}
\centering
\includegraphics[width=\linewidth]{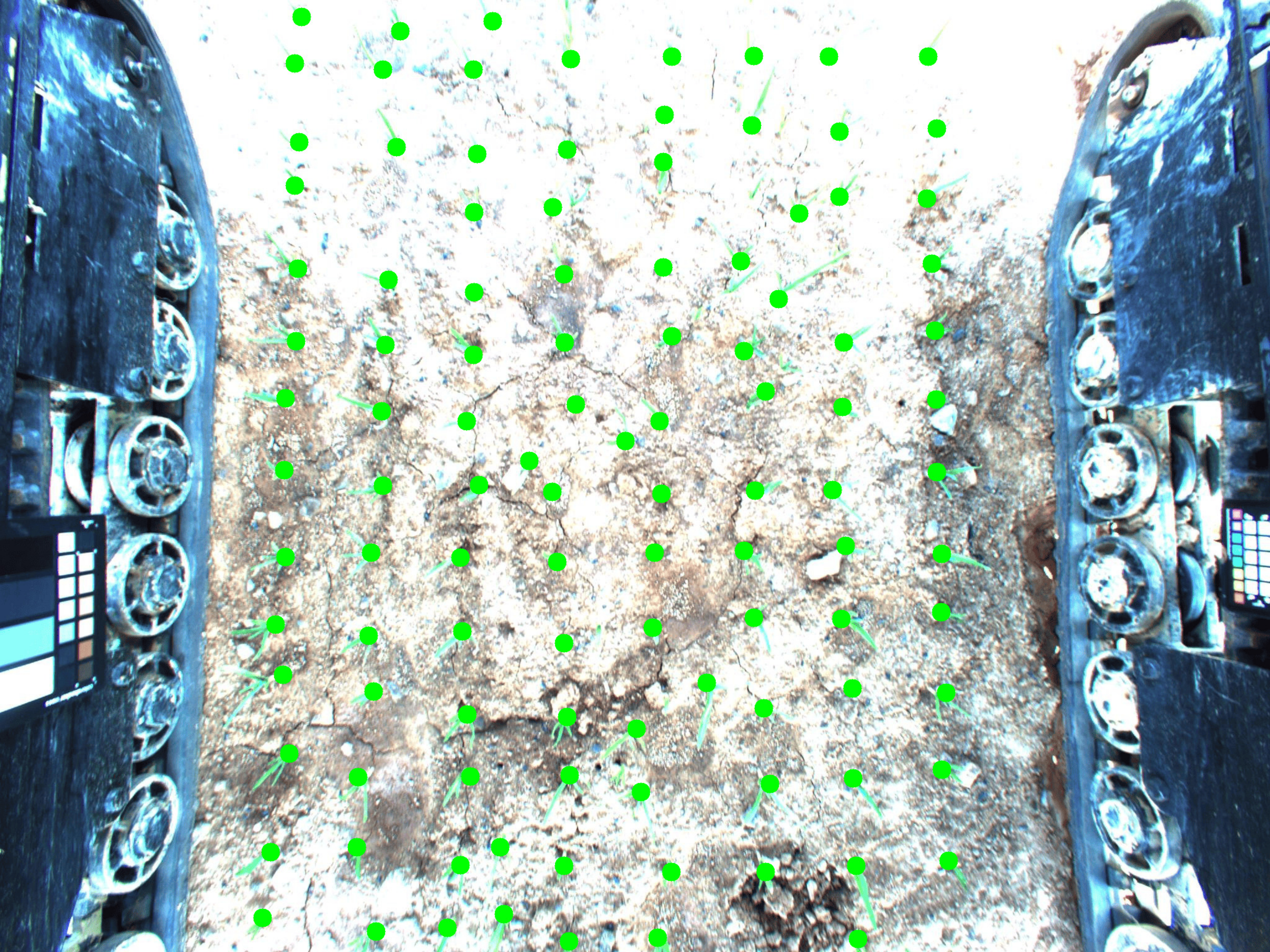}
\end{minipage}
\begin{minipage}{0.25\linewidth}
\centering
\includegraphics[width=\linewidth, height=3.45cm]{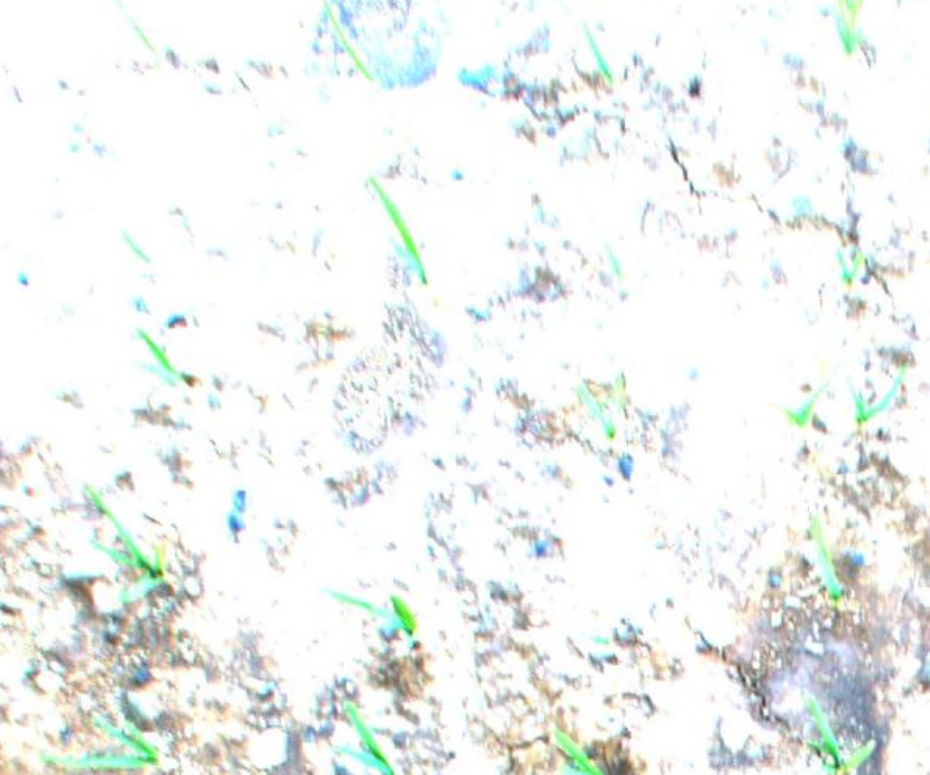}
\end{minipage}
\\

\begin{minipage}{0.25\linewidth}
\centering
\includegraphics[width=\linewidth]{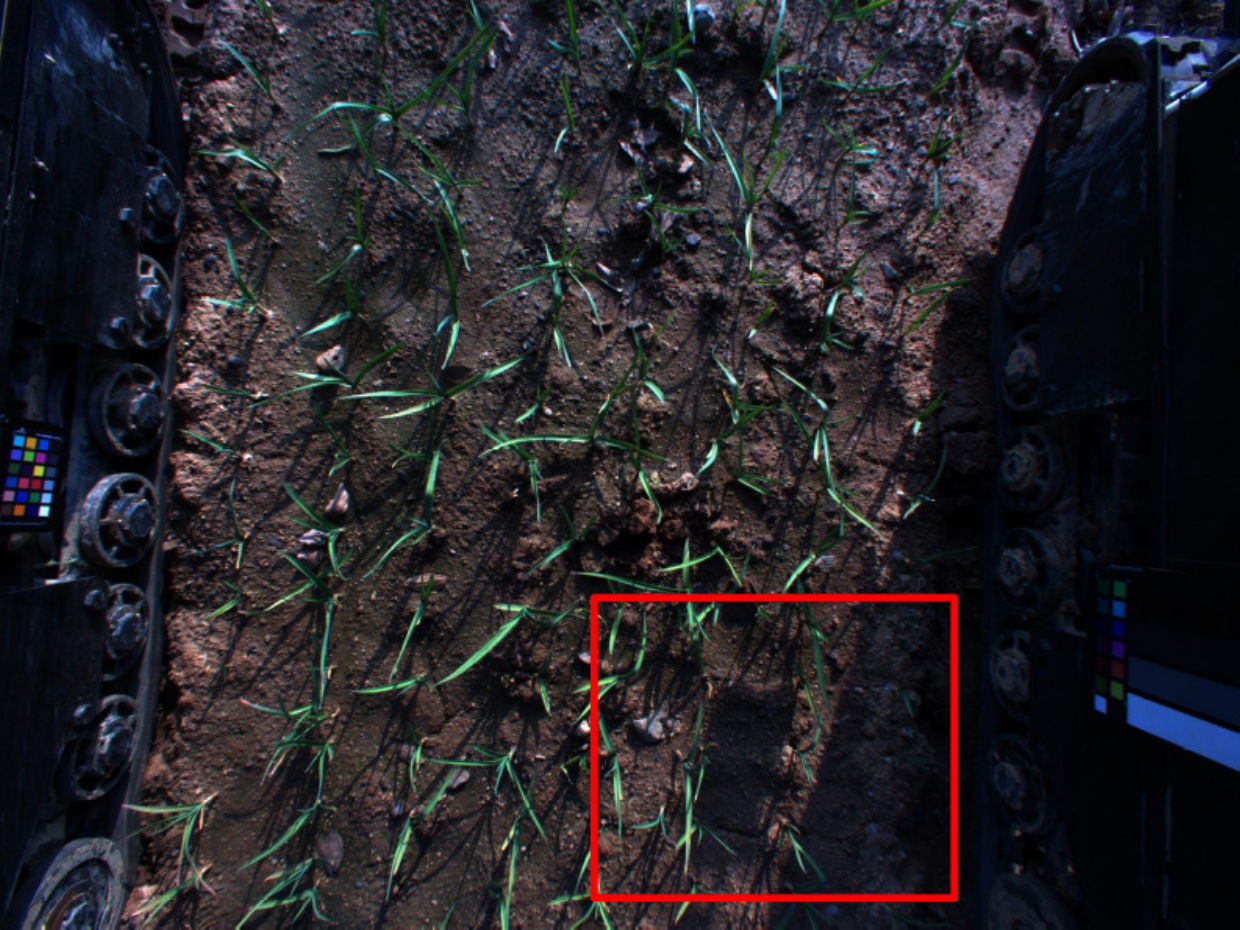}
\end{minipage}
\begin{minipage}{0.25\linewidth}
\centering
\includegraphics[width=\linewidth]{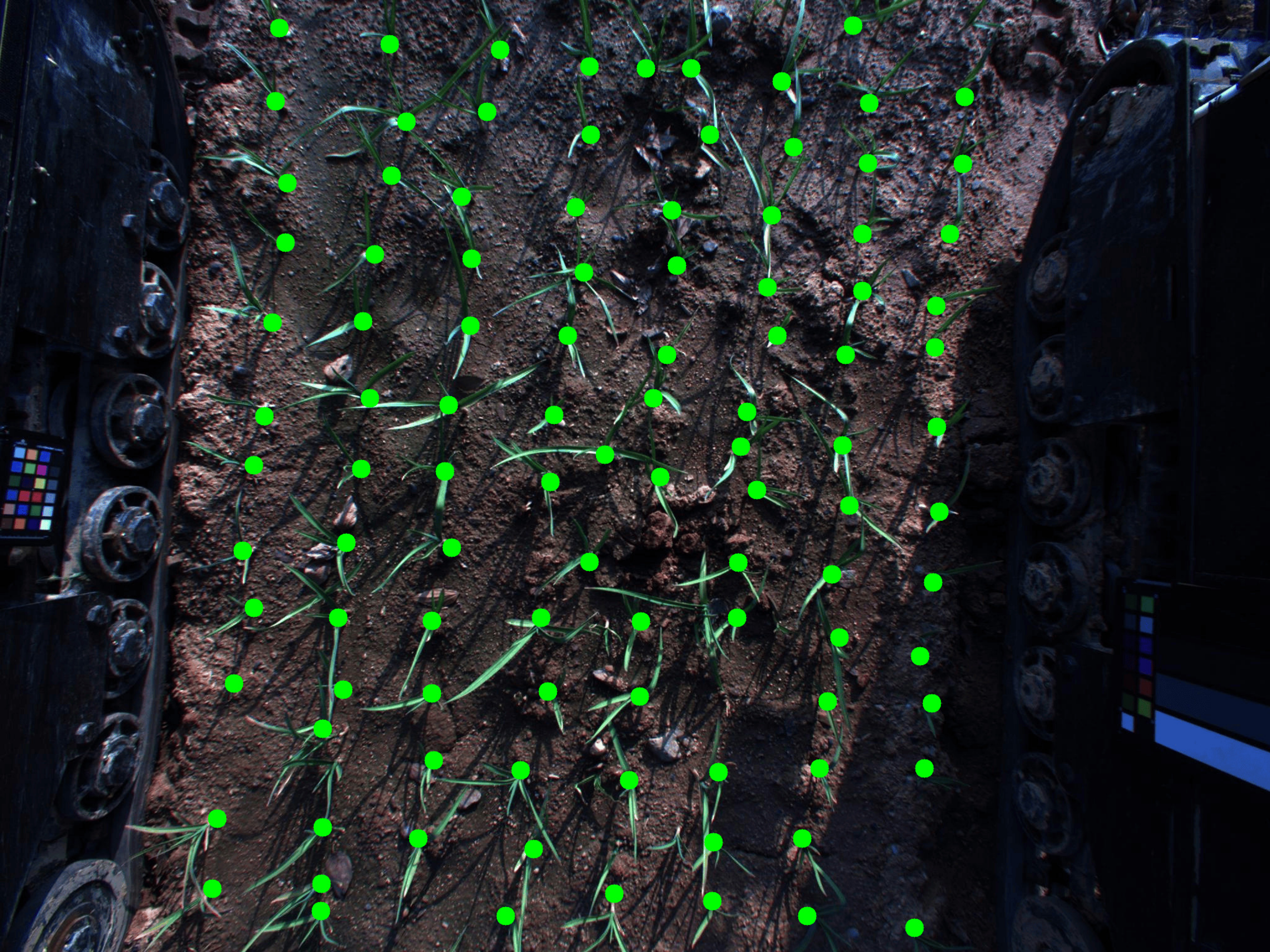}
\end{minipage}
\begin{minipage}{0.25\linewidth}
\centering
\includegraphics[width=\linewidth, height=3.45cm]{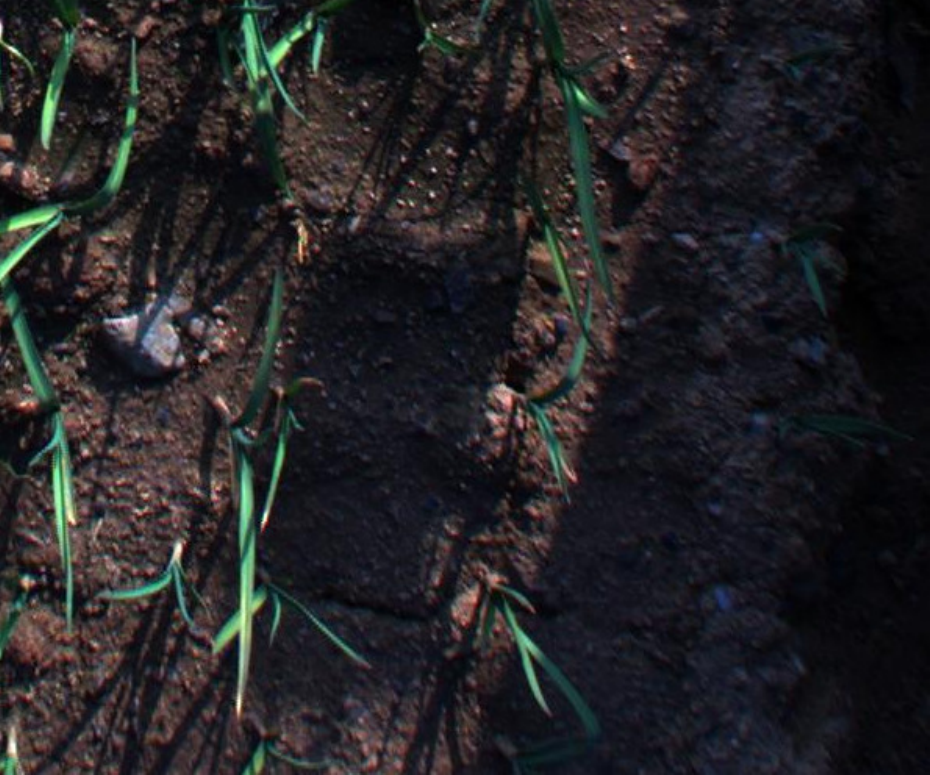}
\end{minipage}
\\

\begin{minipage}{0.25\linewidth}
\centering
\includegraphics[width=\linewidth]{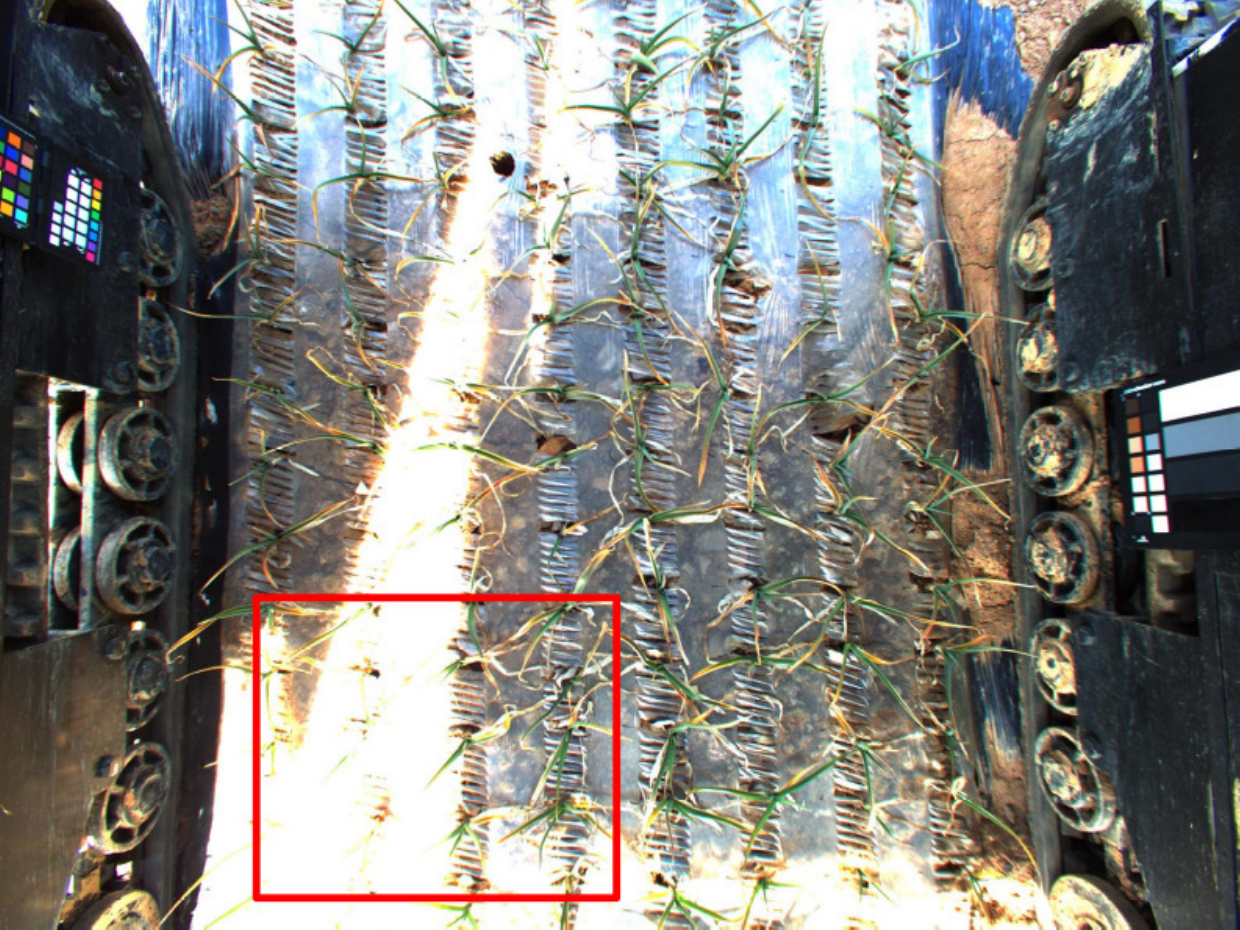}
\end{minipage}
\begin{minipage}{0.25\linewidth}
\centering
\includegraphics[width=\linewidth]{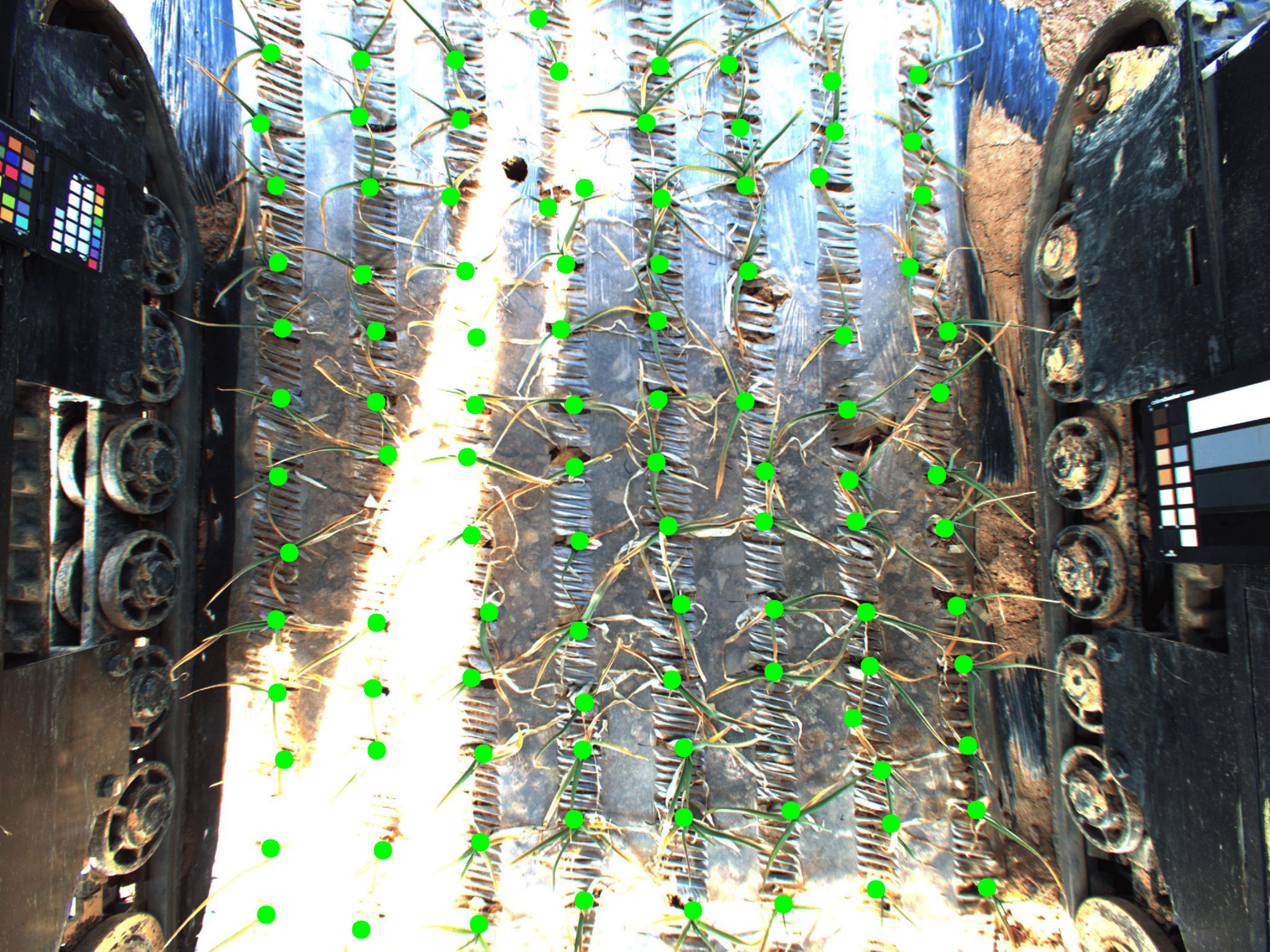}
\end{minipage}
\begin{minipage}{0.25\linewidth}
\centering
\includegraphics[width=\linewidth, height=3.45cm]{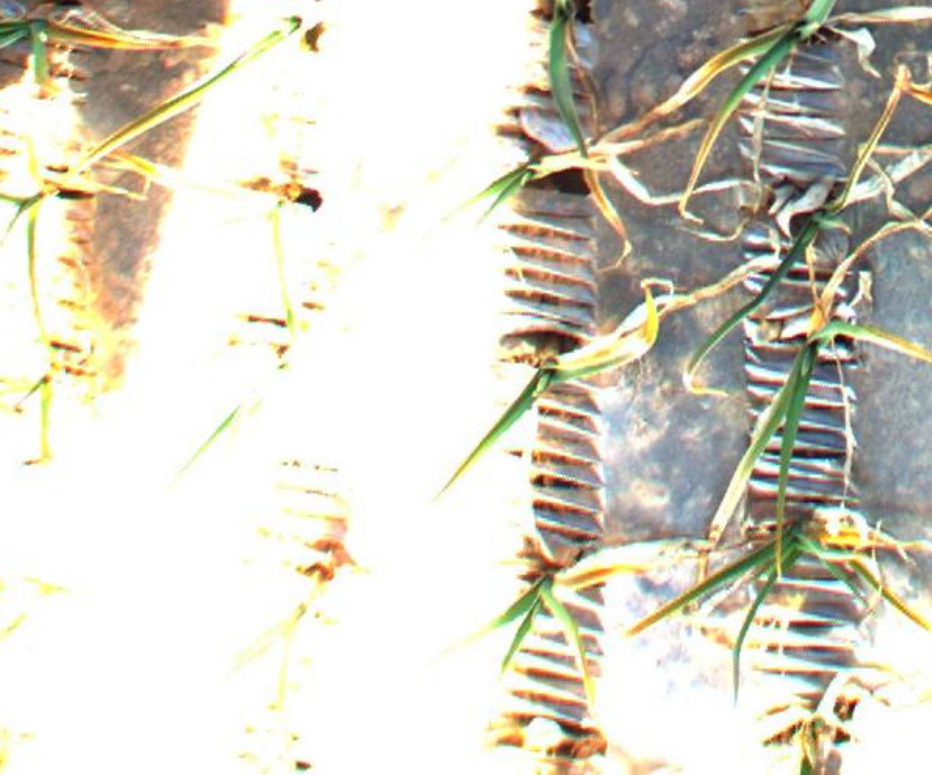}
\end{minipage}
\\

\begin{minipage}{0.25\linewidth}
\centering
\includegraphics[width=\linewidth]{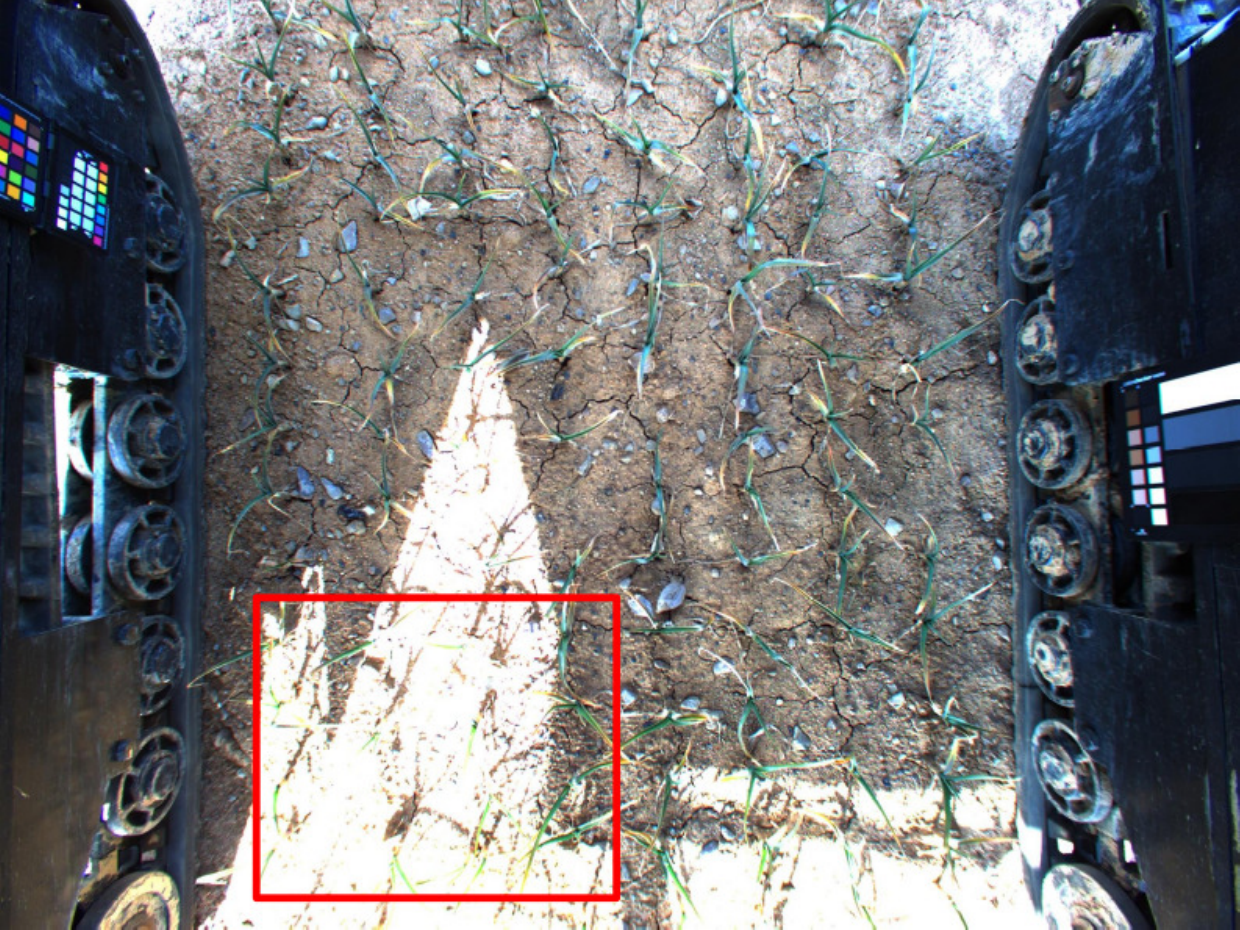}
\end{minipage}
\begin{minipage}{0.25\linewidth}
\centering
\includegraphics[width=\linewidth]{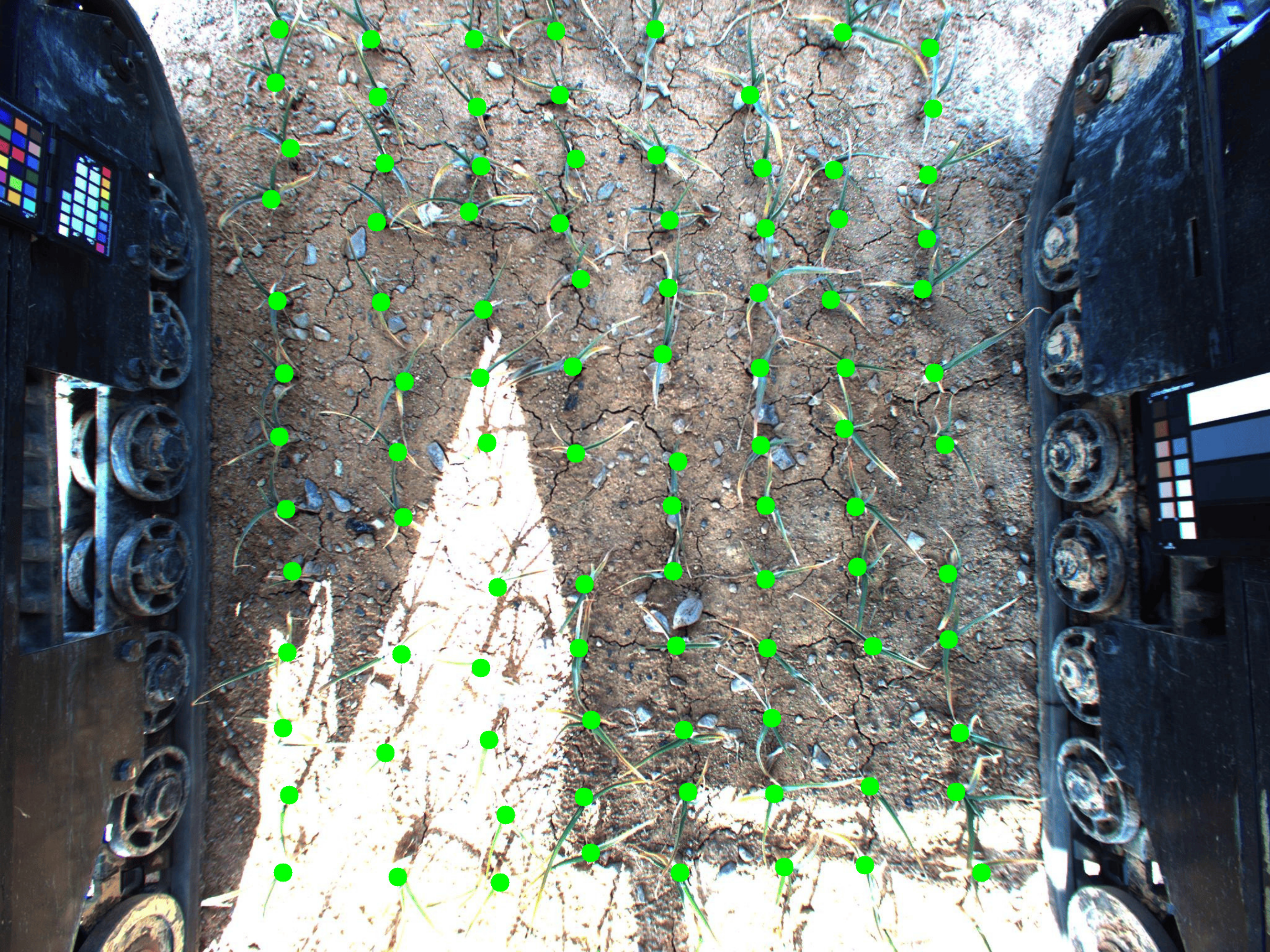}
\end{minipage}
\begin{minipage}{0.25\linewidth}
\centering
\includegraphics[width=\linewidth, height=3.45cm]{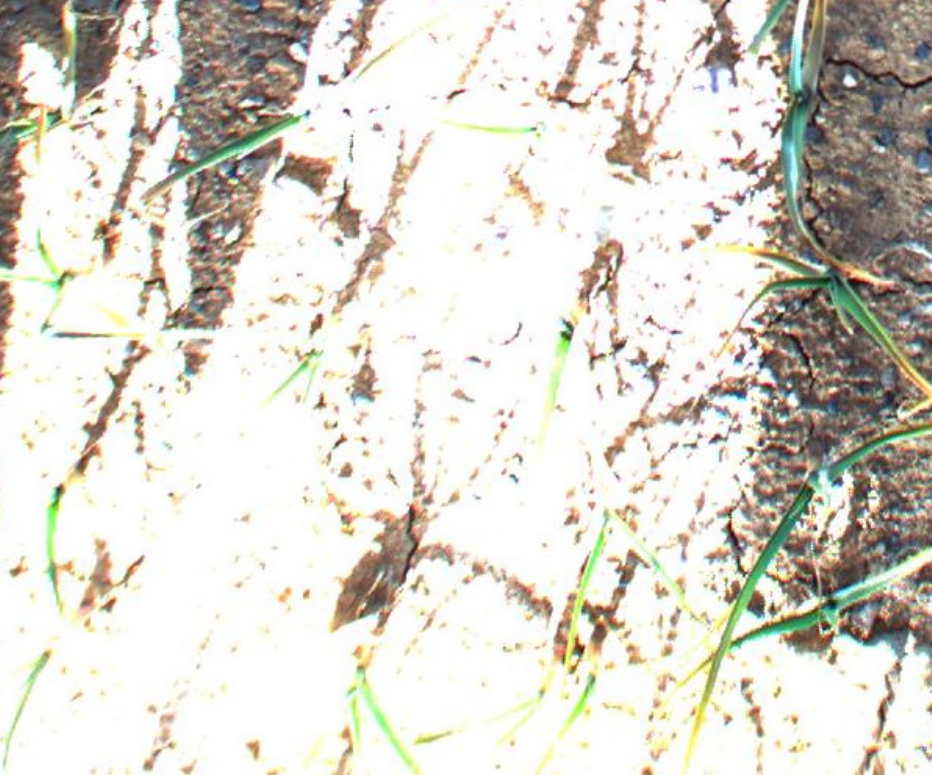}
\end{minipage}
\\

\begin{minipage}{0.25\linewidth}
\centering \small (a) Original image
\end{minipage}
\begin{minipage}{0.25\linewidth}
\centering \small (b) Image with labels
\end{minipage}
\begin{minipage}{0.25\linewidth}
\centering \small (c) Enlarged image
\end{minipage}
\caption{Examples of collected images and their annotations under diverse outdoor illumination conditions. The images include over-exposed regions, under-exposed or shadowed regions, and spatially heterogeneous illumination, which can reduce seedling visibility and increase the difficulty of reliable detection. (a) shows a raw image captured by the monitoring system, and (b) presents the corresponding image with human-annotated labels. (c) provides a zoomed-in view of the region highlighted by the red box in (a). Green dots indicate the locations of garlic seedlings.}
\label{fig:data}
\end{figure*}

Although these approaches have achieved encouraging results, most prior studies assess seedling detection using datasets acquired under relatively stable lighting conditions, such as UAV imagery or greenhouse settings. Consequently, the capability of these methods to handle severe illumination variability in real-world outdoor environments remains insufficiently investigated. In practical field scenarios, particularly when ground-based monitoring platforms are employed, images are frequently captured under highly dynamic lighting conditions caused by direct sunlight, strong shadows, and reflective surfaces. Such illumination variations significantly increase the difficulty of achieving reliable seedling detection.

More specifically, these illumination variations lead to several failure modes in ground-based seedling detection, as illustrated in~\fref{fig:data}. In over-exposed regions, the appearance of small seedlings can be washed out, reducing the color and texture contrast between seedlings and the surrounding soil. In under-exposed or shadowed regions, seedling features may become suppressed or visually confused with background clutter. Moreover, spatially heterogeneous illumination within a single image can cause the same seedling class to exhibit substantially different visual characteristics depending on its location in the frame. These factors may result in missed detections, false positives, and inaccurate localization of seedling centers. Since missing seedling locations are subsequently inferred from the spatial distribution of detected seedlings, such detection errors can further propagate to downstream missing seedling localization. Therefore, the central problem addressed in this study is robust seedling detection under spatially heterogeneous outdoor illumination, while maintaining accurate seedling localization for subsequent field analysis.

To bridge this gap, we design a ground-based field monitoring platform and construct a new dataset for garlic seedling detection collected under real agricultural conditions with uncontrolled outdoor illumination. Using this challenging dataset, we explore adversarial augmentation policy learning to enhance detection robustness, aiming to achieve accurate and dependable seedling detection in practical farming scenarios.

Traditional object detection models often experience substantial performance degradation when operating under difficult illumination conditions~\citep{Hong_2024_YOLA, Du_2024_DAINet}. To alleviate this problem, many studies have introduced image enhancement or preprocessing components into detection frameworks~\citep{Liu_2022_IAYolo, Kalwar_2023_GDIP, XI_2024_DroneDet}. These methods typically attempt to improve the visual quality of input images before performing detection, and the enhancement modules are frequently trained jointly with the detection networks. More recently, research efforts have increasingly focused on learning illumination-robust feature representations that enhance detection performance under varying lighting conditions without explicitly relying on image enhancement procedures~\citep{Cui_2021_MAET, Du_2024_DAINet}.

\begin{figure}[t] 
\begin{minipage}{1\linewidth}
\centerline{\includegraphics[scale=0.6]{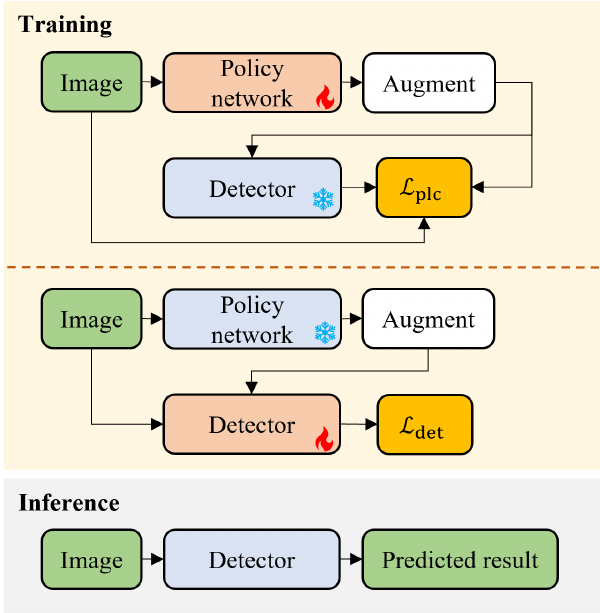}}
\end{minipage}
\caption{Overview of the proposed framework. During training, the policy network and the detector are optimized in an alternating manner. In the policy learning step (top), the policy network is trained while the detector is frozen. The policy network predicts augmentation policies that generate augmented images through the augmentation module, and the policy loss $\mathcal{L}_{\text{plc}}$ is used to update the policy network. In the detector training step (middle), the policy network is frozen and used to generate augmentation policies, while the detector is trained using both original and augmented images by minimizing the detection loss $\mathcal{L}_{\text{det}}$. During inference (bottom), only the detector is used to produce the final prediction without additional modules or computational overhead. Red and blue boxes denote learnable and frozen modules, respectively. Yellow boxes represent loss functions, and green boxes denote inputs and outputs.}
\label{fig:teaser}
\end{figure}

Although these approaches have shown improved robustness in general object detection scenarios, their effectiveness for seedling detection in real agricultural environments remains insufficiently explored. In this study, we address this limitation by introducing an adversarial augmentation policy learning framework specifically designed for seedling detection in ground-based agricultural monitoring systems. The proposed method learns augmentation strategies that intentionally generate challenging variations, thereby enhancing the detector's resilience to adverse visual conditions. To achieve this, the augmentation policy agent and the object detector are jointly optimized so that the detector becomes robust to diverse outdoor illumination variations. The overall workflow of the proposed method is illustrated in Figure~\ref{fig:teaser}.

Our main contributions are summarized as follows:
\begin{itemize}
\item We formulate ground-based garlic seedling detection under severe and spatially heterogeneous outdoor illumination as a robustness problem that directly affects downstream missing seedling localization. To support this study, we construct a new garlic seedling dataset collected under real field conditions using a ground-based monitoring platform.
\item We propose an adversarial augmentation policy learning framework in which the policy network and the detector interact through alternating optimization. The policy network generates input-dependent challenging augmentations based on detector feedback, while the detector is trained using both original and policy-augmented images to learn illumination-robust seedling representations.
\item We integrate the detection loss and a structural penalty into the policy reward so that the adversarial augmentation process increases detection difficulty while preserving the structural consistency of seedling images. This interaction helps prevent the policy from producing unrealistic artifacts and encourages meaningful robustness-oriented augmentations.
\item We demonstrate that the proposed training-time policy learning strategy improves seedling detection and downstream missing seedling localization under challenging illumination conditions, while introducing no additional inference-time computational overhead because only the detector is used during deployment.
\end{itemize}

\section{Related Work}
\subsection{Seedling Detection}
Seedling detection is a fundamental component of precision agriculture, enabling automated crop monitoring, yield estimation, and early-stage field management. Early studies primarily relied on conventional image-processing techniques, such as threshold-based segmentation and color-index-driven methods, due to their simplicity and computational efficiency~\citep{Gennaro_2020_evaluation, Gao_2022_design, CHEN_2022_assimilation, Yuan_2024_rapidly}.

Representative works in this category commonly employed vegetation indices combined with global thresholding. For example, \cite{Gennaro_2020_evaluation} applied Otsu thresholding to UAV imagery to separate seedlings from soil and identified missing plants by detecting canopy gaps. Similarly, \cite{Gao_2022_design} integrated Otsu's method with a majority voting strategy to detect unseeded holes in plug trays using a machine vision system. In outdoor wheat fields, \cite{CHEN_2022_assimilation} utilized the excess-green-minus-excess-red (ExGR) index followed by Otsu thresholding to segment seedlings from soil in UAV images. Likewise, \cite{Yuan_2024_rapidly} adopted the excess green (ExG) index together with a modified Hough transform to identify missing cotton seedlings from aerial imagery.

Despite their effectiveness under controlled conditions, these traditional approaches are highly sensitive to illumination variations, shadows, and changes in soil appearance, as they depend heavily on fixed thresholds and low-level visual features. To address these limitations, recent studies have increasingly adopted deep-learning-based detection frameworks~\citep{Li_2021_high_precision, YAN_2023_machine, Cui_2023_real_time, Wu_2025_novel}.

Several deep-learning-based approaches focus on indoor or greenhouse environments, where lighting conditions are relatively stable~\citep{Li_2021_high_precision, YAN_2023_machine}. For instance, \cite{Li_2021_high_precision} built a dataset of hydroponic lettuce seedlings collected under controlled greenhouse illumination and trained a Faster R-CNN~\citep{Ren_2017_FasterRCNN}-based detector with an enhanced backbone to better exploit high-resolution features. Similarly, \cite{YAN_2023_machine} employed a YOLOv5~\citep{yolov5}-based model to detect missing tomato seedlings in plug trays under uniform LED lighting.

In contrast, several studies have investigated seedling detection in outdoor agricultural environments, which are characterized by more complex and dynamic conditions~\citep{Cui_2023_real_time, Wu_2025_novel}. These methods predominantly rely on UAV imagery. Specifically, \cite{Cui_2023_real_time} proposed a tracking-by-detection framework based on ByteTrack~\citep{Zhang_2022_Bytetrack}, in which missing rice seedlings were inferred from abnormal spatial distances between adjacent detections. To further enhance detection accuracy, \cite{Wu_2025_novel} incorporated transplanter trajectory information and planting geometry to reconstruct expected seedling positions, enabling more accurate identification of missing rice seedlings.

\cite{Anandakrishnan_2024_LiYOLOv9} recently proposed Li-YOLOv9 for precise spatial prediction of rice seedlings from large-scale airborne remote sensing data, emphasizing lightweight and accurate detection for UAV-based monitoring. While this work demonstrates the importance of efficient rice seedling detection in field-scale remote sensing, our study addresses a different setting: close-range ground-based garlic seedling detection under severe spatially heterogeneous illumination. Rather than designing a new lightweight detector, we improve robustness through training-time adversarial augmentation policy learning without increasing inference-time overhead.

Despite these advances, existing studies based on indoor imaging systems~\citep{Li_2021_high_precision, YAN_2023_machine} or UAV platforms~\citep{Cui_2023_real_time, Wu_2025_novel} generally assume relatively uniform illumination conditions. In contrast, ground-based outdoor monitoring systems capture close-range images that often exhibit severe and spatially heterogeneous illumination variations, even within a single frame. To address this challenge, we propose a seedling detection framework specifically designed for ground-based outdoor monitoring. We introduce a new dataset collected under real-world outdoor conditions and develop a training strategy aimed at improving robustness to illumination changes. Furthermore, unlike prior works that primarily focus on rice~\citep{Cui_2023_real_time, Wu_2025_novel}, tomato~\citep{YAN_2023_machine}, or hydroponic lettuce~\citep{Li_2021_high_precision}, our study targets seedling detection in outdoor garlic fields.

\subsection{Object Detection under Challenging Illumination Conditions}
Recent object detection models have demonstrated impressive performance on standard benchmarks under well-controlled lighting conditions~\citep{Zhao2024DETRs, Tian2025YOLOv12, Lin2014COCO}. However, their performance often deteriorates substantially under challenging illumination scenarios, such as low-light or spatially uneven lighting environments~\citep{Hong_2024_YOLA, Du_2024_DAINet, RAL_2022_Bang}. Although numerous low-light image enhancement methods have been developed~\citep{Guo_2020_ZeroDCE, Liu_2021_RUAS, Ma_2022_SCI}, directly applying these techniques as a preprocessing step for object detection has been shown to provide limited benefits or even to degrade detection performance~\citep{Hashmi_2023_FeatEnHancer}.

To overcome this limitation, a line of research has focused on task-driven image enhancement, in which image enhancement is explicitly optimized for object detection. DB-GAN~\citep{Minciullo_2021_DBGAN}, for example, trains a generative adversarial network by jointly optimizing GAN-related losses and detection losses, enabling the generator to produce images tailored to detection tasks. Other studies adopt differentiable image processing pipelines and learn enhancement parameters in an end-to-end manner~\citep{Liu_2022_IAYolo, Kalwar_2023_GDIP, XI_2024_DroneDet}. In particular, Image-Adaptive YOLO~\citep{Liu_2022_IAYolo} jointly trains an object detector and a lightweight CNN to estimate parameters for a differentiable image enhancement module. This framework was further extended by GDIP~\citep{Kalwar_2023_GDIP}, which parallelizes image processing operations, and by DEDet~\citep{XI_2024_DroneDet}, which predicts pixel-wise enhancement parameters.

Unlike methods based on pre-defined image processing pipelines~\citep{Liu_2022_IAYolo, Kalwar_2023_GDIP, XI_2024_DroneDet}, FeatEnHancer~\citep{Hashmi_2023_FeatEnHancer} learns to enhance feature maps using CNNs and jointly optimizes them with task-specific objectives. Similarly, YOLA~\citep{Hong_2024_YOLA} seeks to encode illumination-invariant features by exploiting the Lambertian reflectance assumption~\citep{Shafer_1992_Lambertian}. While these methods achieve improved performance under challenging illumination conditions, they generally require additional processing modules during inference, resulting in increased computational overhead~\citep{Minciullo_2021_DBGAN, Kalwar_2023_GDIP, XI_2024_DroneDet, Hong_2024_YOLA}.

To eliminate additional inference costs, recent studies have explored training-time strategies for learning illumination-invariant representations~\citep{Cui_2021_MAET, Du_2024_DAINet}. MAET~\citep{Cui_2021_MAET} employs an encoder-decoder architecture to learn illumination-invariant features from paired normal-light and synthetically generated low-light images. Similarly, DAI-Net~\citep{Du_2024_DAINet} utilizes a pre-trained Retinex decomposition network~\citep{Chen_2018_RetinexNet} to disentangle illumination-invariant and illumination-related components, relying on synthetic low-light images generated via a dark ISP pipeline.

Inspired by these training-time strategies~\citep{Cui_2021_MAET, Du_2024_DAINet}, we propose a streamlined object detection framework that introduces no additional computational overhead during inference. Moreover, while methods such as DAI-Net~\citep{Du_2024_DAINet} and MAET~\citep{Cui_2021_MAET} depend on pre-trained decomposition networks or traditional image signal processing (ISP) pipelines, the proposed approach does not rely on auxiliary pre-trained models or fixed procedures. Instead, it jointly optimizes an input-dependent adversarial augmentation policy agent together with the detector, enabling the model to adapt more effectively to diverse real-world illumination variations encountered in outdoor environments.

From a deployment-oriented perspective, \cite{bakirci2025performance} recently evaluated lightweight YOLO detectors for real-time monitoring on resource-constrained drone systems, analyzing the trade-offs among accuracy, inference latency, GPU usage, and power consumption. This work highlights that practical monitoring systems should consider not only robustness but also computational efficiency and deployment constraints. In contrast to lightweight detector design studies~\citep{bakirci2025performance, 8014790}, our framework improves robustness through training-time adversarial augmentation and does not introduce additional inference-time overhead, thereby preserving deployment efficiency.

\subsection{Augmentation Policy Learning}
Conventional data augmentation methods rely on fixed, hand-crafted rules that are uniformly applied across datasets, which can be suboptimal for specific tasks or data distributions. To overcome this limitation, learning-based approaches have been proposed to automatically discover effective augmentation policies~\citep{RAL_2022_Bang}.

AutoAugment~\citep{Cubuk_2019_AutoAugment} pioneered this direction by introducing a reinforcement learning framework that searches for augmentation policies through validation performance optimization. However, its two-stage design, which decouples policy search from task network training, incurs substantial computational overhead. To improve efficiency, OHL-Auto-Aug~\citep{Lin_2019_OHL_AutoAug} proposed a bi-level optimization strategy that jointly learns augmentation policies and task networks in an end-to-end manner.

Beyond validation-driven objectives~\citep{Cubuk_2019_AutoAugment, Lin_2019_OHL_AutoAug}, several studies have explored adversarial formulations for augmentation policy learning. Adversarial AutoAugment~\citep{Zhang_2020_Adversarial_Autoaugm}, for instance, trains a policy network to generate challenging augmented samples by maximizing the training loss of the task network. Despite their effectiveness, these methods learn input-agnostic augmentation policies that are applied uniformly to all samples, thereby overlooking class-level or instance-level characteristics~\citep{Cubuk_2019_AutoAugment, Lin_2019_OHL_AutoAug, Zhang_2020_Adversarial_Autoaugm}.

To address this limitation, recent works have investigated class-adaptive or instance-adaptive augmentation strategies~\citep{cheung_2022_adaaug, Miao_2022_InstaAug, Li_2025_AROID}. AdaAug~\citep{cheung_2022_adaaug} learns class-dependent and optionally instance-dependent augmentation policies by minimizing validation loss. InstaAug~\citep{Miao_2022_InstaAug} further advances this paradigm by learning input-conditioned transformation parameters for individual samples based on the task training objective. In a related direction, AROID~\citep{Li_2025_AROID} proposes instance-wise augmentation policies to enhance robustness against adversarial perturbations.

Inspired by recent instance-adaptive augmentation approaches~\citep{cheung_2022_adaaug, Miao_2022_InstaAug, Li_2025_AROID}, we investigate instance-wise augmentation strategies that learn input-conditioned distributions over augmentation operations and their magnitudes. Furthermore, motivated by adversarial augmentation frameworks~\citep{Zhang_2020_Adversarial_Autoaugm, Li_2025_AROID, KANG2024108456}, we incorporate adversarial objectives to further enhance the robustness of the detection model. 

Different from previous methods, we introduce a structural penalty term alongside the adversarial objective to restrict excessive structural distortions in augmented samples. The augmentation policy network is jointly optimized with the seedling detector, enabling the detector to learn more robust representations under challenging visual conditions.

\begin{figure*}[t] 
\begin{minipage}{1\linewidth}
\centerline{\includegraphics[scale=0.6]{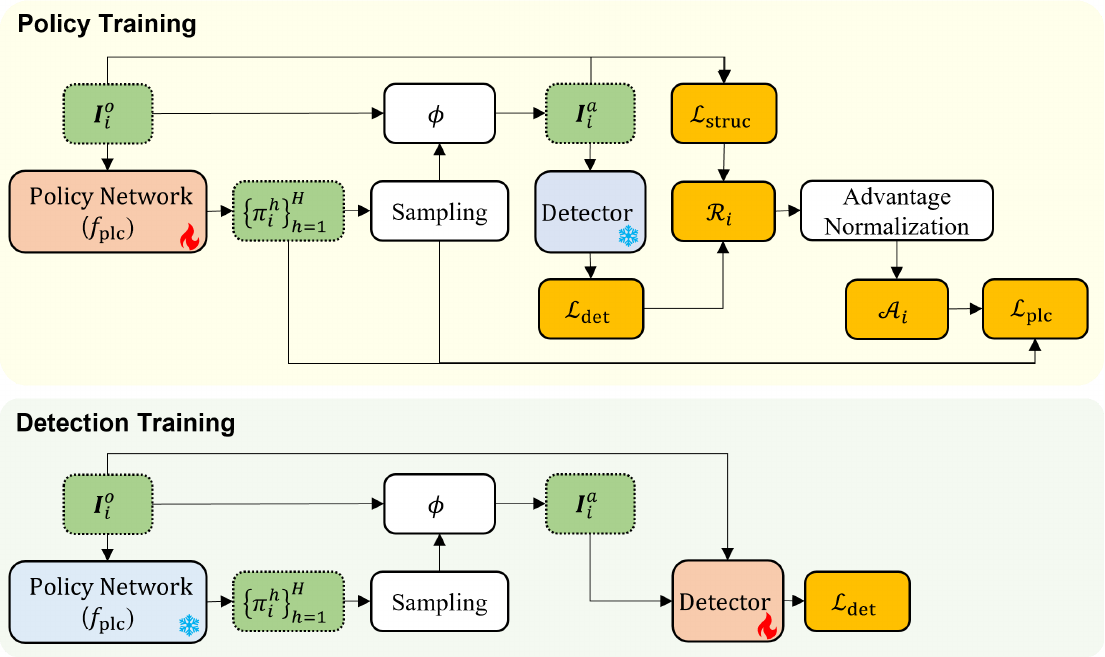}}
\end{minipage}
\caption{Overview of the proposed training framework consisting of policy training and detection training. In the policy training stage (top), the policy network is optimized using a reinforcement learning objective. Given an input image $\mI_i^{o}$, the policy network predicts distributions over augmentation operations. Augmentation actions are sampled and applied through the transformation function $\phi$ to generate an augmented image $\mI_i^{a}$. The augmented image is evaluated by the frozen detector to compute the detection loss $\mathcal{L}_{\text{det}}$. A structural penalty $\mathcal{L}_{\text{struc}}$ is additionally computed to discourage excessive distortions, and the reward $\mathcal{R}_i$ is defined based on these two terms. After advantage normalization, the policy network is updated by minimizing the policy loss $\mathcal{L}_{\text{plc}}$. In the detection training stage (bottom), the policy network is frozen and used to generate augmentation policies for each input image. The detector is then trained using both the original images $\mI_i^{o}$ and the policy-augmented images $\mI_i^{a}$ by minimizing the detection loss $\mathcal{L}_{\text{det}}$. Red and blue boxes denote learnable and frozen modules, respectively. Yellow boxes represent loss functions, and green boxes denote inputs and outputs.}
\label{fig:training_overview}
\end{figure*}

\section{Garlic Seedling Dataset (GSD)}
\label{sec:dataset}
To the best of our knowledge, we present the first dataset specifically designed for garlic seedling analysis. The dataset was collected using a custom-designed, ground-based monitoring system developed to capture close-range images under real-world outdoor conditions.

The monitoring system consists of two main components: a drive module and an image acquisition module. The drive module employs a battery-powered, crawler-based mobility platform, which enables stable operation over unpaved and uneven agricultural terrain. The image acquisition module is equipped with a TRI032S-CC industrial camera housing featuring a Sony IMX265 CMOS sensor with a resolution of 2048$\times$1536. Image capture and data storage are handled by a LattePanda single-board computer connected to the camera.

The image acquisition module is mounted on an adjustable vertical frame, allowing the camera height to be adapted to different crop types and growth stages. To ensure consistent camera alignment, a motorized leveling system compensates for slope variations in the field. Mechanical vibrations are mitigated through shock-absorbing elements installed between the camera sensor and the mobile platform. In addition, removable blackout curtains are used to partially shield the camera from extreme natural illumination. The entire platform is powered by four 48V lithium-polymer batteries, with a DC-DC converter supplying 12V devices, enabling up to 12 hours of continuous operation for both locomotion and image acquisition.

Using this monitoring system, we collected a total of 561 top-down images containing garlic seedlings. Each seedling was annotated with a point label, as illustrated in~Figure\ref{fig:data}. Point-based annotations were selected instead of bounding boxes to reduce annotation time and cost while preserving precise localization information. In total, the dataset contains 55,516 annotated garlic seedlings across all images.

\section{Proposed Method}
We propose a policy-driven augmentation framework that learns to generate adversarial and input-dependent transformations for improving object detection robustness. The core idea is to train a stochastic augmentation policy that produces challenging perturbations while preserving the structural integrity of the input image. The learned policy is jointly optimized with the detection network through an alternating training procedure. Figure~\ref{fig:training_overview} illustrates the overall framework. In the following, we first introduce the augmentation policy agent and its training procedure, and then describe the joint optimization of the policy and detection networks.

\subsection{Augmentation Policy Agent}
\label{sec:stage1}
We introduce an augmentation policy agent that learns to generate adversarial and input-dependent augmentation strategies. The agent is implemented as a stochastic policy network that predicts transformation policies conditioned on the input image. The policy is trained to produce challenging augmented samples that increase the detection loss while preserving the structural integrity of the original image. We first describe the training procedure of the policy network, followed by the design of the augmentation space.

\subsubsection{Training Policy Agent}
To implement an adaptive and input-conditioned augmentation strategy, we employ a stochastic policy network $f_{\text{plc}}$ with parameters $\theta_{\text{plc}}$, which predicts an augmentation policy for each input image. The network consists of a shared convolutional backbone followed by four parallel linear heads. Each head is responsible for a specific augmentation category, namely \textit{light adjustment}, \textit{texture detail}, \textit{drop out}, and \textit{region selection}. From its corresponding augmentation sub-space, each head samples an operation, as summarized in Table~\ref{tab:aug_space}.

\begin{table*}[!t]
\centering
\caption{
Augmentation space used by the policy network. Each policy head selects one operation from its corresponding sub-space. Some operations are associated with a magnitude that controls the strength of the transformation.
}
\label{tab:aug_space}
\footnotesize
\begin{tabular}{
>{\centering}m{0.12\textwidth} >{\centering}m{0.07\textwidth}|
>{\centering}m{0.15\textwidth} >{\centering}m{0.08\textwidth}|
>{\centering}m{0.12\textwidth} >{\centering}m{0.07\textwidth}|
>{\centering}m{0.12\textwidth} >{\centering\arraybackslash}m{0.07\textwidth}}
\toprule
\multicolumn{2}{c|}{Region Selection} & \multicolumn{2}{c|}{Light Adjustment} & \multicolumn{2}{c|}{Texture Detail} & \multicolumn{2}{c}{Drop Out}  \\
\midrule
Operation        & Magnitude  & Operation      & Magnitude & Operation           & Magnitude & Operation           & Magnitude \\
\midrule
Full Image            & --  & Identity    & --         & Identity       & --    & Identity               & --     \\
Over-exposed   & -- & Brightness  & [0.1, 1.0] & Sharpness      & 2.0   & Erasing                & --   \\
Under-exposed  & -- & Contrast    & [0.1, 1.0] & Gaussian Blur  & 3.0   & Pixelate               & --     \\
Normal-exposed & -- & Gamma       & [0.1, 1.0] & Glass Blur     & 0.7   & Elastic Transform & --     \\
                    &     & Hue         & [0.1, 1.0] & Motion Blur    & 0.5   &  &   \\
                    &     & Saturation  & [0.1, 1.0] & Median Blur    & --    &  &   \\
                    &     & Solarize    & [0.1, 1.0] & Gaussian Noise & --    &  &   \\
                    &     & Posterize   & [0.1, 1.0] & ISO Noise      & --    &  &   \\
                    &     & Invert      & --         &                &       &  &   \\
                    &     & Histogram Equalization   & --             &       &  &                     &   \\
                    &     & Log Transform       & --             &       &  &                     &   \\
\bottomrule
\end{tabular}
\end{table*}

The four-head design follows the functional decomposition of the augmentation space rather than an arbitrary choice of model complexity. Since each head is tied to a specific augmentation sub-space, changing the number of heads would require redefining the augmentation space itself, such as merging or splitting transformation categories. Therefore, the number of heads is not treated as an independent architectural hyperparameter. This design provides a task-driven balance between policy expressiveness, training stability, and training cost.

Each prediction head $h$ outputs a categorical distribution $\pi_i^{h}$ over candidate operation-magnitude pairs within its corresponding augmentation category. These distributions $\{\pi_i^{h}\}_{h=1}^{H}$ represent the policy’s preference for selecting each transformation, where $H$ denotes the number of augmentation heads. During sampling, an action $s_i^{h}$ is independently drawn from each distribution (\ie $s_i^{h} \sim \pi_i^{h}$).

The actions sampled from the heads corresponding to \textit{light adjustment}, \textit{texture detail}, and \textit{drop out} are sequentially applied to the input image $\mI_i^o$, producing an initially augmented image $\hat{\mI}_i^a$. The action sampled from the \textit{region selection} head generates a spatial mask $\mM_i^a$ that determines the region where the sampled perturbations are applied. The final augmented image $\mI_i^{a}$ is then obtained by blending the original image $\mI_i^o$ and the initially augmented image $\hat{\mI}_i^a$ using the spatial mask $\mM_i^a$:
\begin{equation}
\label{eq:blend}
\mI_i^{a} := \mM_i^a \cdot \hat{\mI}_i^a + (1 - \mM_i^a) \cdot \mI_i^o,
\end{equation}
where $\cdot$ denotes element-wise multiplication. This mechanism allows the policy to apply adversarial perturbations to spatially localized regions.

To optimize the policy network $f_{\text{plc}}$, we employ the REINFORCE algorithm~\citep{Williams_REINFORCE_1992b}, a Monte Carlo policy gradient method for training stochastic policies. REINFORCE is adopted due to its simplicity and effectiveness in optimizing stochastic policies, following prior augmentation-policy learning approaches~\citep{Lin_2019_OHL_AutoAug, Li_2025_AROID}. Unlike these methods, however, we directly use the detection loss as the reward signal, since the policy is designed to adversarially perturb the detection network, as illustrated in Algorithm~\ref{alg:policy}.

Specifically, given an input image $\mI_i^o$, the policy network outputs the distributions $\{\pi_i^{h}\}_{h=1}^{H}$ over candidate operation-magnitude pairs. The policy is then optimized by performing $T$ rollouts. At each rollout $t$, a set of transformation parameters $\calS_i^{(t)}=\{s_i^{h,(t)}\}_{h=1}^{H}$ is sampled from the stochastic policy $\{\pi_i^{h}\}_{h=1}^{H}$, where $s_i^{h,(t)} \sim \pi_i^{h}$. The sampled transformations are applied through the augmentation function $\phi$, generating a policy-driven augmented image $\mI_i^{a,(t)} = \phi(\mI_i^o, \calS_i^{(t)})$, as illustrated in Figure~\ref{fig:training_overview}.

\begin{algorithm}[t]
\caption{Augmentation Policy Optimization}
\label{alg:policy}
\textbf{Input:} mini-batch $\{\mI_i^o\}_{i=1}^{B}$, policy parameters $\theta_{\text{plc}}$, frozen detector parameters $\theta_{\text{det}}$, number of rollouts $T$, number of heads $H$, learning rate $\alpha_{\text{plc}}$ \\
\textbf{Output:} updated policy parameters $\theta_{\text{plc}}$ \\

\For{$i = 1$ \KwTo $B$}{
    $\{\pi_i^{h}\}_{h=1}^{H} = f_{\text{plc}}(\mI_i^o; \theta_{\text{plc}})$ \\
    \For{$t = 1$ \KwTo $T$}{
        Sample $s_i^{h,(t)} \sim \pi_i^{h}, \ \forall h$ \\
        $\calS_i^{(t)} = \{ s_i^{h,(t)} \}_{h=1}^{H}$ \\
        $\mI_i^{a,(t)} = \phi(\mI_i^o, \calS_i^{(t)})$ \\
        $\mathcal{R}_i^{(t)} = \mathcal{L}_{\text{det}}(\mI_i^{a,(t)}; \theta_{\text{det}}) - \beta_{\text{struc}} \mathcal{L}_{\text{struc}}(\mI_i^o, \mI_i^{a,(t)})$ \\
        $\log \mathcal{P}_i^{(t)} = \sum_{h=1}^{H} \log \pi_i^{h}\!\left(s_i^{h,(t)}\right)$
    }
$\mu_{\mathcal{R}} = \frac{1}{T} \sum_{t=1}^{T} \mathcal{R}_i^{(t)}$, \\ 
$\sigma_{\mathcal{R}} = \sqrt{\frac{1}{T} \sum_{t=1}^{T} (\mathcal{R}_i^{(t)} - \mu_{\mathcal{R}})^2}$ \\
    \For{$t = 1$ \KwTo $T$}{
$\mathcal{A}_i^{(t)} = \frac{\mathcal{R}_i^{(t)} - \mu_{\mathcal{R}}}{\sigma_{\mathcal{R}} + \epsilon}$ \\
}
}
$\mathcal{L}_{\text{plc}} = -\frac{1}{BT} \sum_{i=1}^{B} \sum_{t=1}^{T} \log \mathcal{P}_i^{(t)} \, \mathcal{A}_i^{(t)}$ \\
$\theta_{\text{plc}} \leftarrow \theta_{\text{plc}} - \alpha_{\text{plc}} \nabla_{\theta_{\text{plc}}} \mathcal{L}_{\text{plc}}$ \\
\end{algorithm}

The policy is trained to generate adversarial augmentations that are challenging for the detector. However, increasing the detection loss alone does not necessarily guarantee improved robustness, since unconstrained adversarial augmentations may introduce unrealistic artifacts. Thus, the detection loss is used as a proxy for identifying challenging training samples, and it is explicitly regularized by a structural penalty. Specifically, the reward signal $\mathcal{R}_i^{(t)}$ is defined as:
\begin{equation}
\mathcal{R}_i^{(t)} := \mathcal{L}_{\text{det}}(\mI_i^{a,(t)}; \theta_{\text{det}}) - \beta_{\text{struc}} \mathcal{L}_{\text{struc}}(\mI_i^o, \mI_i^{a,(t)}),
\label{eq:reward}
\end{equation}
where $\theta_{\text{det}}$ denotes the parameters of the object detector. The structural penalty is defined as $\mathcal{L}_{\text{struc}} = 1 - \text{SSIM}(\mI_i^o, \mI_i^{a,(t)})$, which discourages excessive structural distortion and helps preserve the spatial layout of the original image. The hyperparameter $\beta_{\text{struc}}$ controls the trade-off between adversarial perturbation and structural preservation. This formulation encourages the policy to generate challenging yet structurally consistent augmentations, rather than simply maximizing the detection loss.

SSIM is adopted because the purpose of the structural penalty is to preserve perceptual and spatial consistency rather than enforce strict pixel-wise similarity. In the proposed framework, illumination-related changes are intentionally encouraged to improve robustness; therefore, pixel-level penalties such as L1 may overly restrict valid photometric variations. In contrast, SSIM evaluates similarity based on luminance, contrast, and structural information, allowing moderate illumination changes while penalizing augmentations that substantially distort the local structure of seedlings.

To stabilize training, we apply advantage normalization to the rewards~\citep{PPO2017}. The normalized advantage $\mathcal{A}_i^{(t)}$ is computed as:
\begin{equation}
\mathcal{A}_i^{(t)} := \frac{\mathcal{R}_i^{(t)} - \mu_{\mathcal{R}}}{\sigma_{\mathcal{R}} + \epsilon},
\label{eq:advantage}
\end{equation}
where $\mu_{\mathcal{R}}$ and $\sigma_{\mathcal{R}}$ denote the mean and standard deviation of the rewards computed over all $T$ rollout samples. Here, $T$ denotes the number of rollouts per input image. The constant $\epsilon$ is a small value added for numerical stability.

Since REINFORCE-based optimization relies on sampled augmentation actions, the estimated policy gradient may have high variance. In our experiments, we did not observe severe training instability such as divergence or collapse of the policy network, although the raw rollout rewards fluctuated across sampled augmentation policies. To mitigate this issue, advantage normalization is applied within the $T$ rollouts for each input image, which reduces reward-scale variation and limits the influence of a small number of high-loss augmentations on the policy update. In addition, periodic policy updates with the frozen detector reduce the non-stationarity of the reward signal, while the structural penalty discourages degenerate augmentations with unrealistic distortions.

The REINFORCE surrogate objective $\mathcal{L}_{\text{plc}}$ is defined as the expected log-probability of the sampled augmentations weighted by their corresponding advantages. The policy gradient is estimated as:
\begin{equation}
    \nabla_{\theta_{\text{plc}}} \mathcal{L}_{\text{plc}} :=
    - \frac{1}{B T} \sum_{i=1}^{B} \sum_{t=1}^{T} \sum_{h=1}^{H}
    \nabla_{\theta_{\text{plc}}} \log \pi_i^{h}\!\left(s_i^{h,(t)}\right)
    \mathcal{A}_i^{(t)},
\end{equation}
where $\pi_i^{h}\!\left(s_i^{h,(t)}\right)$ denotes the probability assigned by the $h$-th policy head to the sampled augmentation action $s_i^{h,(t)}$ for the input image $\mI_i^o$. This objective encourages the policy to increase the likelihood of augmentation strategies that induce larger detection errors while preserving the semantic structure of the image.

The policy network is trained by minimizing $\mathcal{L}_{\text{plc}}$, yielding the optimal policy parameters $\theta_{\text{plc}}^* = \arg\min_{\theta_{\text{plc}}} \mathcal{L}_{\text{plc}}$. The policy network is optimized using stochastic gradient descent over sampled mini-batches.

\subsubsection{Augmentation Space}
\label{sec:action_space}
We design a structured augmentation space that enables the policy network to generate diverse illumination variations and adversarial perturbations. As summarized in Table~\ref{tab:aug_space}, the augmentation space is organized into four sub-spaces: \textit{light adjustment}, \textit{texture detail}, \textit{drop out}, and \textit{region selection}.

The \textit{region selection} sub-space determines the spatial region where the sampled perturbations are applied. It contains four options: \textit{full image}, \textit{over-exposed}, \textit{under-exposed}, and \textit{normal-exposed}. Exposure regions are estimated using the approach proposed in~\citep{Guo_2010_exposure_map}, which operates in the CIELAB color space. Specifically, regions with high luminance but low saturation are categorized as over-exposed, whereas regions with low luminance and low saturation are categorized as under-exposed. All remaining areas are considered normal-exposed regions. The transformations sampled from the remaining sub-spaces are applied only within the selected region, as defined in Eq.~\ref{eq:blend}.

The \textit{light adjustment} sub-space contains photometric transformations that simulate diverse illumination conditions, including brightness, contrast, gamma, hue, saturation, solarize, and posterize. For parameterized operations, the magnitude is uniformly discretized into 10 levels within the range [0.1, 1.0]. Other operations are applied using their default settings.

The \textit{texture detail} sub-space consists of transformations that perturb local texture characteristics, including various blur and noise operations.

The \textit{drop out} sub-space includes operations such as erasing, pixelate, and elastic transformation, which intentionally remove or distort local structural information to simulate localized corruption.

\subsection{Joint Training of Policy Agent and Detection Network}
\label{sec:joint_training}
The proposed framework employs an alternating training scheme that interleaves detector updates with periodic optimization of the policy agent, as illustrated in Figure~\ref{fig:training_overview} and Algorithm~\ref{alg:highlevel}. Throughout the $N_{\text{epoch}}$ training epochs, the detection network is optimized at every iteration. In contrast, the policy network $f_{\text{plc}}$ is updated every $K$ iterations using Algorithm~\ref{alg:policy}.

At each iteration, the stochastic policy network predicts a set of per-head action distributions $\{\pi_i^{h}\}_{h=1}^{H}$ for a mini-batch of original images $\mI_i^o$. Augmentation actions $\calS_i$ are then sampled from these distributions and applied through the augmentation function $\phi$, producing augmented images $\mI_i^a = \phi(\mI_i^o, \calS_i)$. The detection network $f_{\text{det}}$ is then optimized using the detection loss $\mathcal{L}_{\text{det}}$ on both the original and augmented images.

During this step, the detection parameters $\theta_{\text{det}}$ are updated via gradient descent, while the policy parameters $\theta_{\text{plc}}$ remain fixed except at scheduled policy update iterations.

This alternating optimization continues for $N_{\text{epoch}}$ epochs, resulting in the joint optimization of the detection network $f_{\text{det}}$ and the augmentation policy network $f_{\text{plc}}$.

\begin{algorithm}[t]
\caption{Joint Training Procedure}
\label{alg:highlevel}
\textbf{Input:} number of epochs $N_{\text{epoch}}$, number of iterations per epoch $N_{\text{iter}}$, policy update interval $K$, dataset $\mathcal{D}$, mini-batch size $B$, learning rate $\alpha_{\text{det}}$ \\
\textbf{Output:} detector parameters $\theta_{\text{det}}$, policy parameters $\theta_{\text{plc}}$ \\
$n \leftarrow 0$ \\
\For{$e = 1$ \KwTo $N_{\text{epoch}}$}{
    \For{$t = 1$ \KwTo $N_{\text{iter}}$}{
        Sample mini-batch $\{\mI_i^{o}\}_{i=1}^{B} \subset \mathcal{D}$ \\
        $\{\pi_i^h\}_{i=1,h=1}^{B,H}= f_{\text{plc}}(\{\mI_i^o\}_{i=1}^{B}; \theta_{\text{plc}})$ \\
        Sample $s_i^h \sim \pi_i^h \quad \forall b, \forall h$ \\
        $\calS_i = \{s_i^h\}_{h=1}^{H} \quad \forall b$ \\
        $\mI_i^{a} = \phi(\mI_i^{o}, \calS_i)$ \\
        $\theta_{\text{det}} \leftarrow \theta_{\text{det}} - \alpha_{\text{det}}
        \nabla_{\theta_{\text{det}}}
        \mathcal{L}_{\text{det}}\!\left(\{\mI_i^{o}, \mI_i^{a}\}_{i=1}^{B}; \theta_{\text{det}}
        \right)$
        $n \leftarrow n + 1$ \\
        \If{$n \bmod K = 0$}{
        Update $\theta_{\text{plc}}$ using Algorithm~\ref{alg:policy} \\
        }
    }
}
\end{algorithm}

\subsection{Missing Seedling Localization}
\label{sec:missing_loc}
In~\sref{sec:joint_training}, the detector is trained to identify only existing seedlings, without explicitly learning a missing-seedling class. This design choice is motivated by the severe class imbalance caused by the relatively small number of missing instances, which may negatively affect detection performance, as well as the high annotation cost associated with labeling missing positions.

Given the detected seedling positions, we analyze their spatial distribution to identify missing seedling locations. The overall procedure consists of two main stages: (1) robust crop row fitting and (2) missing seedling localization.

\vspace{1mm}
\noindent\textbf{Robust Crop Row Fitting}.
Let $\calP=\{\vp_1, \vp_2, \dots, \vp_N\}$ denote the set of detected seedling centers, where $\vp_i=(x_i, y_i)$. To robustly estimate the crop rows from these points, we employ the Random Sample Consensus (RANSAC) algorithm.

Given the regular planting structure of garlic fields and the camera viewpoint, the $k$-th crop row $L_k$ is modeled as a linear function:
\begin{equation}
x = \gamma_k y + \eta_k,
\end{equation}
where $\gamma_k$ and $\eta_k$ denote the slope and intercept of the row, respectively. RANSAC iteratively estimates these parameters by maximizing the number of inlier points associated with each row.

After estimating the row models, each point $\vp_i$ is assigned to the nearest row $L_k$ according to its perpendicular Euclidean distance. This process partitions the detected seedlings into distinct crop rows. For each row, the assigned seedlings are sorted according to their $y_i$ coordinates, and the interval $\Delta_y$ between adjacent seedlings is computed.

\vspace{1mm}
\noindent\textbf{Missing Seedling Localization}.
To identify missing seedlings, we define a threshold $\tau_{\Delta}$ as a scalar multiple of the median interval $\tilde{\Delta}_y$ computed over the entire test set. A gap is classified as missing if the corresponding interval $\Delta_y$ exceeds $\tau_{\Delta}$.

Once a missing gap is detected, the number of missing seedlings $n_{\text{miss}}$ is estimated as:
\begin{equation}
n_{\text{miss}} = \mathrm{round}\left( \frac{\Delta_y}{\tilde{\Delta}_y} \right) - 1.
\end{equation}

Finally, the position of the $j$-th missing seedling $m_j = (\hat{x}_j, \hat{y}_j)$, where $1 \le j \le n_{\text{miss}}$, is estimated as follows. The $y$-coordinates $\hat{y}_j$ are obtained by uniformly dividing the gap, and the corresponding $x$-coordinates $\hat{x}_j$ are computed by projecting $\hat{y}_j$ onto the estimated crop row $L_k$:
\begin{align}
\hat{y}_j &= y_i + \Delta_y \cdot \frac{j}{n_{\text{miss}} + 1}, \\
\hat{x}_j &= \gamma_k \hat{y}_j + \eta_k,
\end{align}
where $y_i$ denotes the $y$-coordinate of the seedling preceding the gap.

This projection strategy based on the estimated crop row is adopted instead of relying only on adjacent seedling positions, as it better accounts for local deviations and provides more geometrically consistent localization.

\begin{table*}[t]
\centering
\caption{Quantitative comparison of seedling detection and missing seedling localization performance with prior methods using YOLOv3~\citep{Redmon_2018_yolov3} under the extreme-brightness stress-test split.}
\label{tab:detection_results}
\begin{minipage}{1\linewidth}
\centering
\begin{tabular}{>{\centering}m{0.3\textwidth} | >{\centering}m{0.1\textwidth} | >{\centering}m{0.1\textwidth} >{\centering}m{0.1\textwidth} >{\centering\arraybackslash}m{0.1\textwidth}}
\toprule
\multirow{2}{*}{Method} & Seedling & \multicolumn{3}{c}{Missing Seedling} \\
\cmidrule(lr){2-2} \cmidrule(lr){3-5}
 & AP$_{50}$ & Precision & Recall & F1-score \\
\midrule
Baseline & 90.7 & 70.2 & \textbf{60.6} & 65.0 \\
FeatEnHancer~\citep{Hashmi_2023_FeatEnHancer} & 91.3 & 71.0 & 58.9 & 64.4 \\
DAI-Net~\citep{Du_2024_DAINet} & 91.4 & 70.3 & 59.9 & 64.7 \\
YOLA~\citep{Hong_2024_YOLA} & 91.3 & 70.1 & 56.2 & 62.4 \\
\textbf{Ours} & \textbf{91.6} & \textbf{75.0} & \textbf{60.6} & \textbf{67.0} \\
\bottomrule
\end{tabular}
\end{minipage}
\end{table*}

\section{Experiments and Results}
\subsection{Experimental Setting}
\label{sec:experimental_setting}
\noindent \textbf{Dataset}.
As described in~\sref{sec:dataset}, the proposed garlic seedling dataset comprises 561 original images with a total of 55,516 annotated seedlings. To reduce annotation costs, seedling locations are labeled using point annotations, as detailed in~\sref{sec:dataset} and illustrated in~\fref{fig:data}. For detector training, each annotated point is converted into a square bounding box centered at the labeled location. The side length of the square is set to approximately 67\% of the average inter-seedling hole distance, as shown in~\fref{fig:result}.

Each original image has a resolution of 2048$\times$1536. To accommodate the spatial scale of seedlings and the input size of the backbone network, each image is divided into four sub-images of size 1024$\times$768. Sub-images that do not contain any seedling annotations are discarded, resulting in 2,234 sub-images used for training and evaluation.

To evaluate robustness under challenging illumination conditions, we first construct an extreme-brightness split. Specifically, the top 10\% and bottom 10\% of images ranked by average brightness are selected to form the test set, and the remaining images are used to construct the training and validation sets. This brightness-based split is intended as a robustness-oriented stress test rather than a fully representative evaluation of the overall field-image distribution. Since this setting may over-emphasize extreme illumination cases, we additionally construct a brightness-balanced split. In this split, images are divided into dark, medium, and bright groups according to their average brightness, and the test set is constructed by sampling the same number of images from each group. This complementary split includes both challenging and moderate illumination conditions, enabling a more balanced assessment across different brightness levels.

The detailed dataset statistics differ between the two evaluation settings. For the extreme-brightness split, the dataset is partitioned into 1,563 training images, 224 validation images, and 447 testing images, containing 39,127, 5,465, and 10,924 annotated seedlings, respectively. For the brightness-balanced split, we construct a separate partition so that the test set contains balanced samples from dark, medium, and bright images. This split contains 1,565 training images, 223 validation images, and 446 testing images, with 39,078, 5,614, and 10,824 annotated seedlings in the corresponding sets, respectively.

Nevertheless, we acknowledge that the scene-level diversity may still be limited compared with large-scale object detection benchmarks, particularly in terms of field scenes, soil conditions, and acquisition environments. In this study, we focus on evaluating the proposed framework on a task-specific garlic seedling dataset collected using a ground-based monitoring platform under real outdoor illumination conditions. We considered cross-dataset evaluation; however, to the best of our knowledge, there is no publicly available dataset that provides a directly comparable setting for ground-based outdoor seedling detection with compatible annotations and evaluation protocols. Existing seedling datasets often differ substantially in imaging platform, acquisition environment, and task definition. For example, many datasets are collected using UAV imagery or greenhouse/indoor imaging systems, whereas our study focuses on close-range ground-based seedling detection and missing seedling localization under outdoor illumination variations. These differences make direct quantitative comparison difficult and may lead to an unfair or misleading evaluation.

\vspace{1mm}
\noindent\textbf{Evaluation metrics.}
For seedling detection, we report Average Precision (AP) computed at an Intersection-over-Union (IoU) threshold of 0.5 between predicted and ground-truth bounding boxes. AP is calculated from the precision-recall curve following the standard object detection evaluation protocol.

For missing seedling localization, we report precision, recall, and F1-score. Since missing seedlings are inferred as point locations rather than detected with bounding boxes, IoU-based metrics are not applicable. Instead, we adopt a Euclidean distance-based matching criterion. A predicted missing location is considered a true positive if its Euclidean distance to the nearest ground-truth missing location is smaller than a predefined threshold $d_{\text{th}}$. Predictions without a corresponding ground-truth match are counted as false positives, while unmatched ground-truth missing locations are treated as false negatives. Each ground-truth missing location is matched with at most one prediction.

\vspace{1mm}
\noindent\textbf{Policy Network}.
The policy network $f_{\text{plc}}$ adopts the Pre-activation ResNet-18~\citep{He_PRN18_2016} with four parallel heads corresponding to the four augmentation categories. The network is trained using SGD with an initial learning rate of 0.001 and a momentum of 0.9. We set the number of rollouts to $T = 10$ and the policy update interval to $K = 6$.

\vspace{1mm}
\noindent\textbf{Detection Network.}
Following DAI-Net~\citep{Du_2024_DAINet}, FeatEnHancer~\citep{Hashmi_2023_FeatEnHancer}, and YOLA~\citep{Hong_2024_YOLA}, we adopt YOLOv3~\citep{Redmon_2018_yolov3} as the main detection backbone to ensure a fair and controlled comparison with existing illumination-robust baseline methods under the same detector architecture. Since the primary objective of this study is to evaluate the effectiveness of the proposed adversarial augmentation policy learning strategy, we keep the detector backbone fixed in the main comparison to isolate the contribution of the proposed training strategy from the influence of different detector architectures. We note that the purpose of this study is not to claim YOLOv3 as the most advanced detector, but to demonstrate the effectiveness of the proposed policy learning framework under a controlled backbone setting. The detector is randomly initialized and trained for 50 epochs with a learning rate of 0.001. All experiments are conducted on an NVIDIA A5000 GPU.

For a fair comparison, all compared methods, including DAI-Net and YOLA, were retrained on the proposed garlic seedling dataset using the same training, validation, and test splits. We used the same YOLOv3 detector backbone, input resolution, annotation conversion strategy, number of training epochs, and evaluation protocol across all methods, except for method-specific modules required by each baseline. To account for method-specific optimization characteristics, the learning rate of each method was selected using the same validation-set-based tuning protocol. For baseline methods with additional components, the corresponding modules were trained jointly with the detector following their original formulations, while keeping the common detector training settings consistent. The missing seedling localization procedure described in Section 4.3 was also applied equally to the detection outputs of all methods.

\subsection{Result}
\label{sec:result}
Table~\ref{tab:detection_results} presents a quantitative comparison of seedling detection and missing seedling localization performance between the proposed method and existing approaches, including FeatEnHancer~\citep{Hashmi_2023_FeatEnHancer}, DAI-Net~\citep{Du_2024_DAINet}, and YOLA~\citep{Hong_2024_YOLA}. The "Baseline" corresponds to training the detector using only the original training images without additional modules or augmentation strategies. Boldface denotes the best-performing results.

For seedling detection, the proposed method achieves the best performance with an $\text{AP}_{50}$ of 91.6\%, outperforming the baseline by an absolute margin of 0.9\% and exceeding the previous best-performing method, DAI-Net~\citep{Du_2024_DAINet}, by 0.2\%. These results indicate that the proposed adversarial augmentation policy effectively improves the robustness of the detector under challenging illumination conditions.

For missing seedling localization, we apply the algorithm described in Section~\ref{sec:missing_loc} to the detection results of all methods. The proposed method achieves the highest precision and F1-score among the compared approaches, reaching 75.0\% precision and an F1-score of 67.0\%. Compared with the baseline, the proposed method improves precision by 4.8\% while maintaining the same recall (60.6\%). This improvement indicates that the proposed method significantly reduces false positive predictions when identifying missing seedlings.

It should be noted that the missing seedling localization performance depends on the quality of the preceding seedling detection results. In our framework, missing seedlings are not directly detected as a separate object class, but are inferred from the spatial distribution of detected existing seedlings through crop-row fitting and gap analysis. For a fair comparison, all compared methods use the same missing seedling localization procedure. Thus, the observed improvement in missing seedling localization mainly reflects the improved quality and consistency of the seedling detection results, rather than a fundamentally superior localization algorithm.

Although the AP$_{50}$ improvement is relatively modest, the result should be interpreted in the context of the target application. The proposed method is primarily designed to improve robustness under challenging illumination conditions, where detection failures are more likely to occur, rather than to substantially improve performance under already favorable imaging conditions. Since the baseline detector already achieves a high AP$_{50}$, large absolute gains are difficult to obtain. More importantly, the proposed method improves missing seedling precision by 4.8 percentage points and F1-score by 2.0 percentage points compared with the baseline while maintaining the same recall. This indicates that the proposed method reduces false positive missing-seedling predictions, which can help avoid unnecessary field inspection or replanting operations. In addition, the proposed framework does not introduce additional computational overhead during inference, since only the detector is used after training.

It is also worth noting that feature-enhancement-based detectors~\citep{Hashmi_2023_FeatEnHancer, Du_2024_DAINet, Hong_2024_YOLA} consistently improve seedling detection accuracy compared with the baseline model. However, their improvements in missing seedling localization remain limited. In contrast, the proposed method improves both detection accuracy and missing seedling localization performance, demonstrating that the learned augmentation policy enables the detector to produce more reliable seedling predictions for downstream spatial analysis.

\begin{table}[t]
\centering
\caption{Additional quantitative comparison using YOLOv3~\citep{Redmon_2018_yolov3} under the brightness-balanced split, where the test set contains balanced samples from dark, medium, and bright images.}
\label{tab:balanced_split_results}
\begin{minipage}{1\linewidth}
\centering
\begin{tabular}{>{\centering}m{0.18\textwidth} | >{\centering}m{0.14\textwidth} | >{\centering}m{0.14\textwidth} >{\centering}m{0.13\textwidth} >{\centering\arraybackslash}m{0.16\textwidth}}
\toprule
\multirow{2}{*}{Method} & Seedling & \multicolumn{3}{c}{Missing Seedling} \\
\cmidrule(lr){2-2} \cmidrule(lr){3-5}
 & AP$_{50}$ & Precision & Recall & F1-score \\
\midrule
Baseline & 91.2 & 74.7 & 66.5 & 70.4 \\
\textbf{Ours} & \textbf{91.8} & \textbf{75.1} & \textbf{67.3} & \textbf{71.0} \\
\bottomrule
\end{tabular}
\end{minipage}
\end{table}

Table~\ref{tab:balanced_split_results} presents the additional evaluation results under the brightness-balanced split. In this setting, the baseline detector achieves an AP$_{50}$ of 91.2\%, while the proposed method improves AP$_{50}$ to 91.8\%. For missing seedling localization, the proposed method improves precision from 74.7\% to 75.1\%, recall from 66.5\% to 67.3\%, and F1-score from 70.4\% to 71.0\%. These results demonstrate that the proposed method also provides consistent improvement under a more balanced brightness distribution. The overall performance gain is smaller than that under the extreme-brightness split, which is expected because the baseline detector performs more reliably under moderate illumination conditions. Therefore, the brightness-balanced evaluation complements the extreme-brightness stress test and indicates that the proposed method is effective under extreme illumination while also remaining beneficial under more balanced field-image conditions.

We additionally evaluate the proposed policy learning strategy using TOOD~\citep{Feng_2021_tood}, a more recent one-stage detector adopted in the YOLA framework~\citep{Hong_2024_YOLA}. This experiment is intended to examine whether the proposed framework can provide robustness benefits beyond the YOLOv3 backbone used in the main comparison. As shown in Table~\ref{tab:tood_results}, the proposed method consistently improves performance over the TOOD baseline. Specifically, AP$_{50}$ increases from 92.0\% to 92.4\%. For missing seedling localization, precision improves from 60.9\% to 63.3\%, recall from 58.9\% to 60.3\%, and F1-score from 59.0\% to 61.8\%. These results suggest that the proposed adversarial augmentation policy learning strategy can also improve robustness when applied to a more recent detector architecture, although a more comprehensive evaluation across a wider range of detector architectures remains future work.

\begin{table}[t]
\centering
\caption{Additional evaluation using TOOD~\citep{Feng_2021_tood} as a more recent one-stage detector backbone.}
\label{tab:tood_results}
\begin{minipage}{1\linewidth}
\centering
\begin{tabular}{>{\centering}m{0.18\textwidth} | >{\centering}m{0.14\textwidth} | >{\centering}m{0.14\textwidth} >{\centering}m{0.13\textwidth} >{\centering\arraybackslash}m{0.16\textwidth}}
\toprule
\multirow{2}{*}{Method} & Seedling & \multicolumn{3}{c}{Missing Seedling} \\
\cmidrule(lr){2-2} \cmidrule(lr){3-5}
& AP$_{50}$ & Precision & Recall & F1-score \\
\midrule
Baseline & 92.0 & 60.9 & 58.9 & 59.0 \\
\textbf{Ours} & \textbf{92.4} & \textbf{63.3} & \textbf{60.3} & \textbf{61.8} \\
\bottomrule
\end{tabular}
\end{minipage}
\end{table}

\begin{figure*}[!t]
\centering
\begin{minipage}{0.24\linewidth}
\centering
\includegraphics[width=\linewidth]{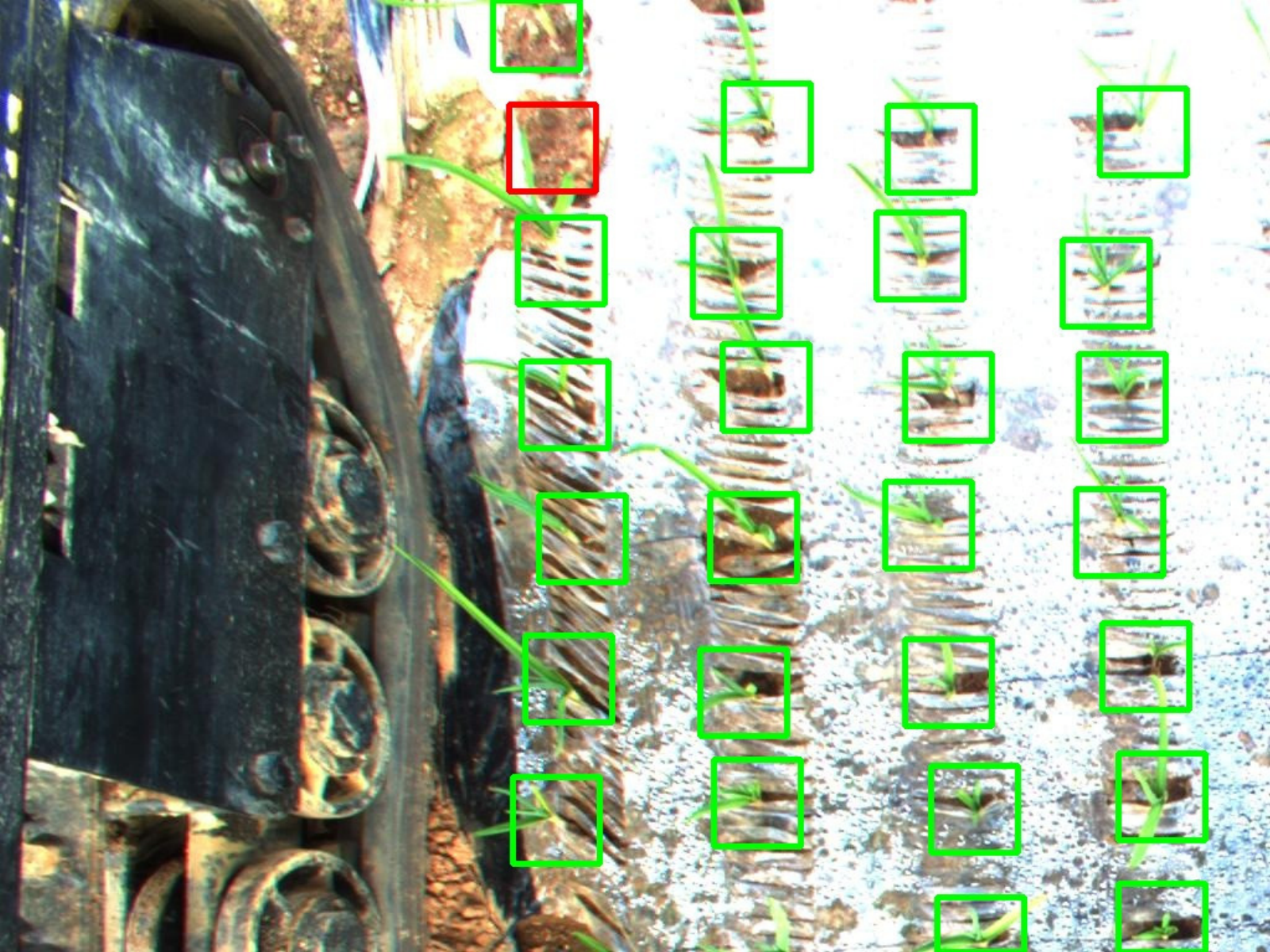}
\end{minipage}
\begin{minipage}{0.24\linewidth}
\centering
\includegraphics[width=\linewidth]{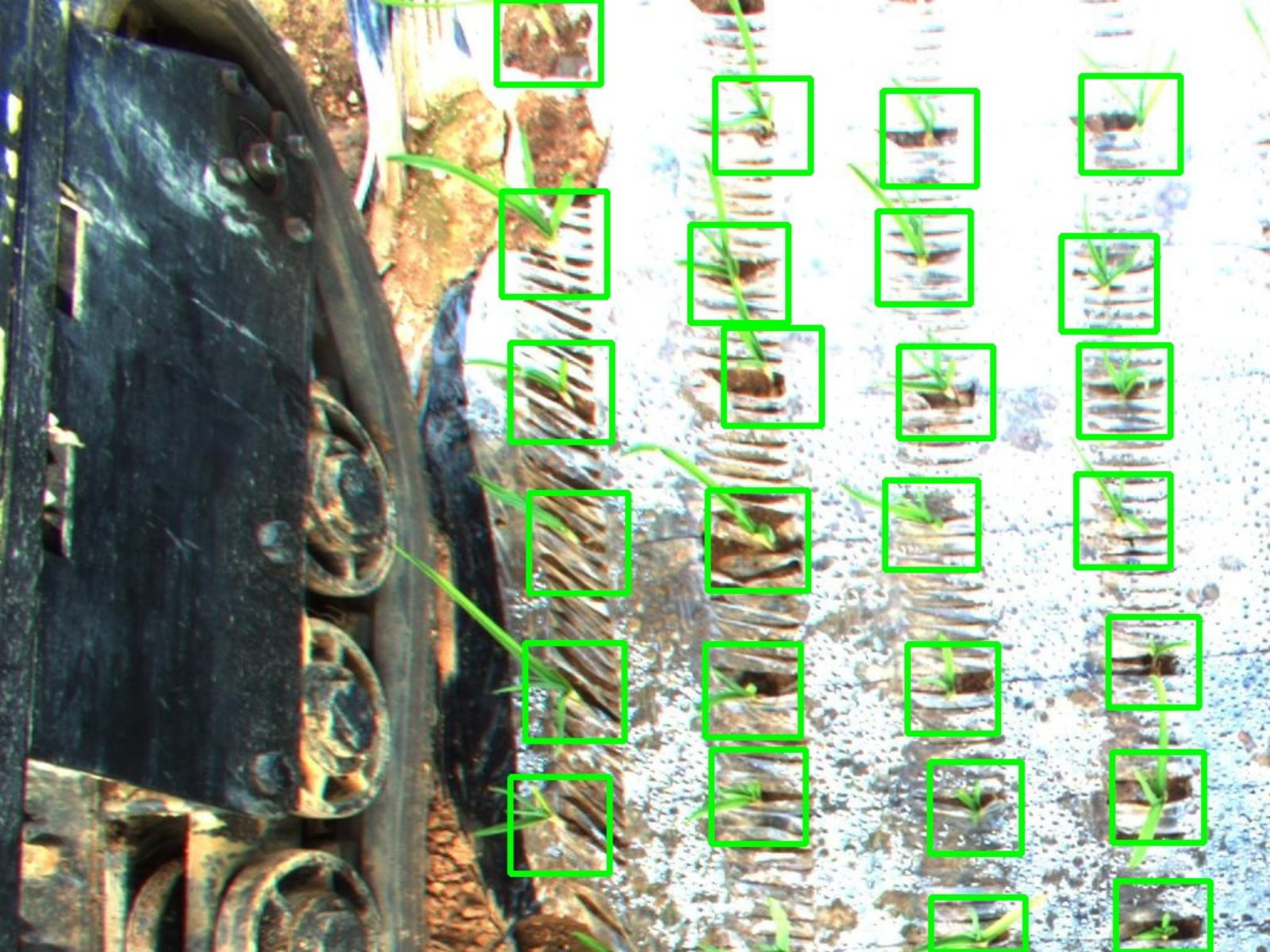}
\end{minipage}
\begin{minipage}{0.24\linewidth}
\centering
\includegraphics[width=\linewidth]{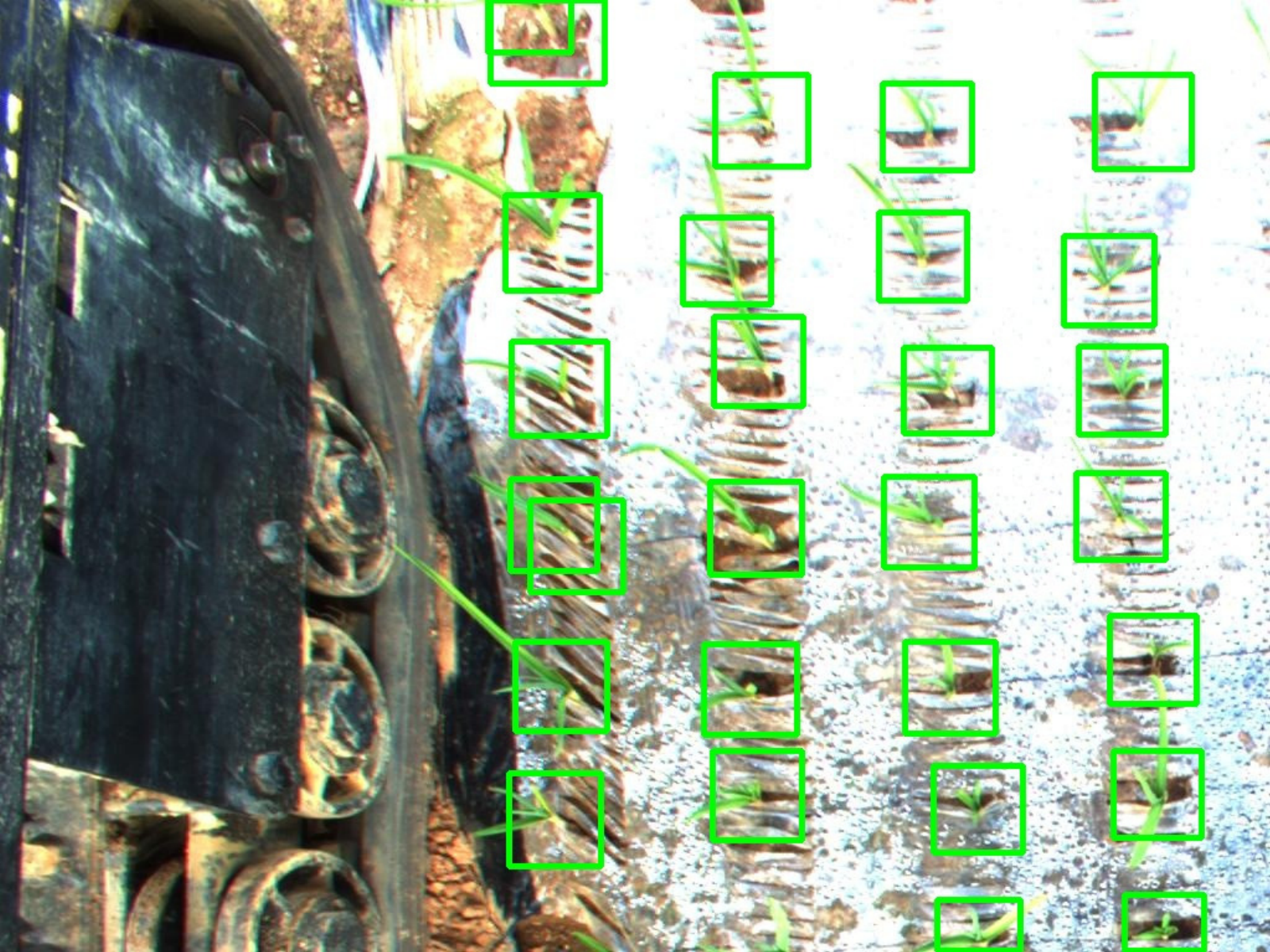}
\end{minipage}
\begin{minipage}{0.24\linewidth}
\centering
\includegraphics[width=\linewidth]{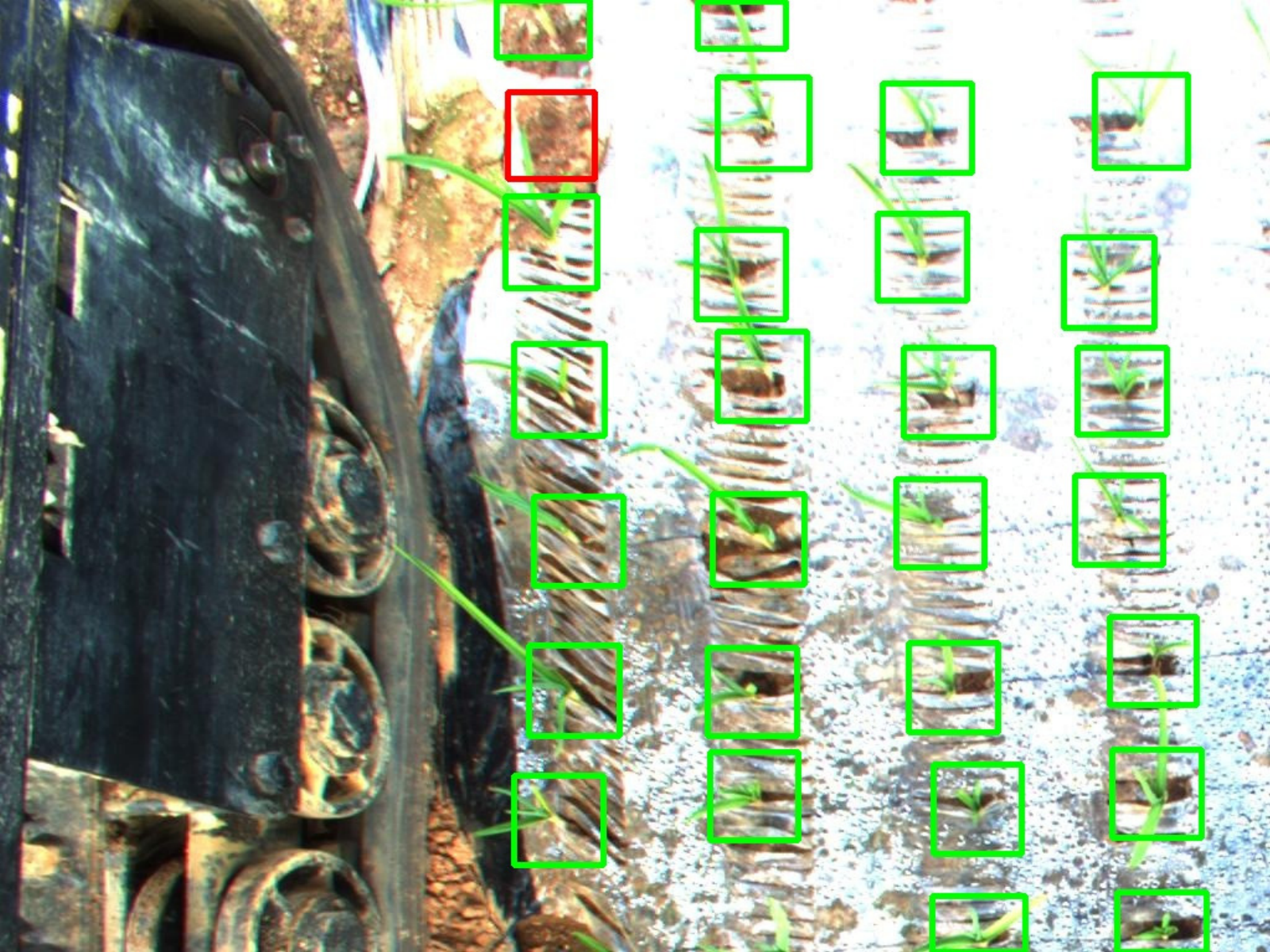}
\end{minipage}
\\
\vspace{0.05cm}

\begin{minipage}{0.24\linewidth}
\centering
\includegraphics[width=\linewidth]{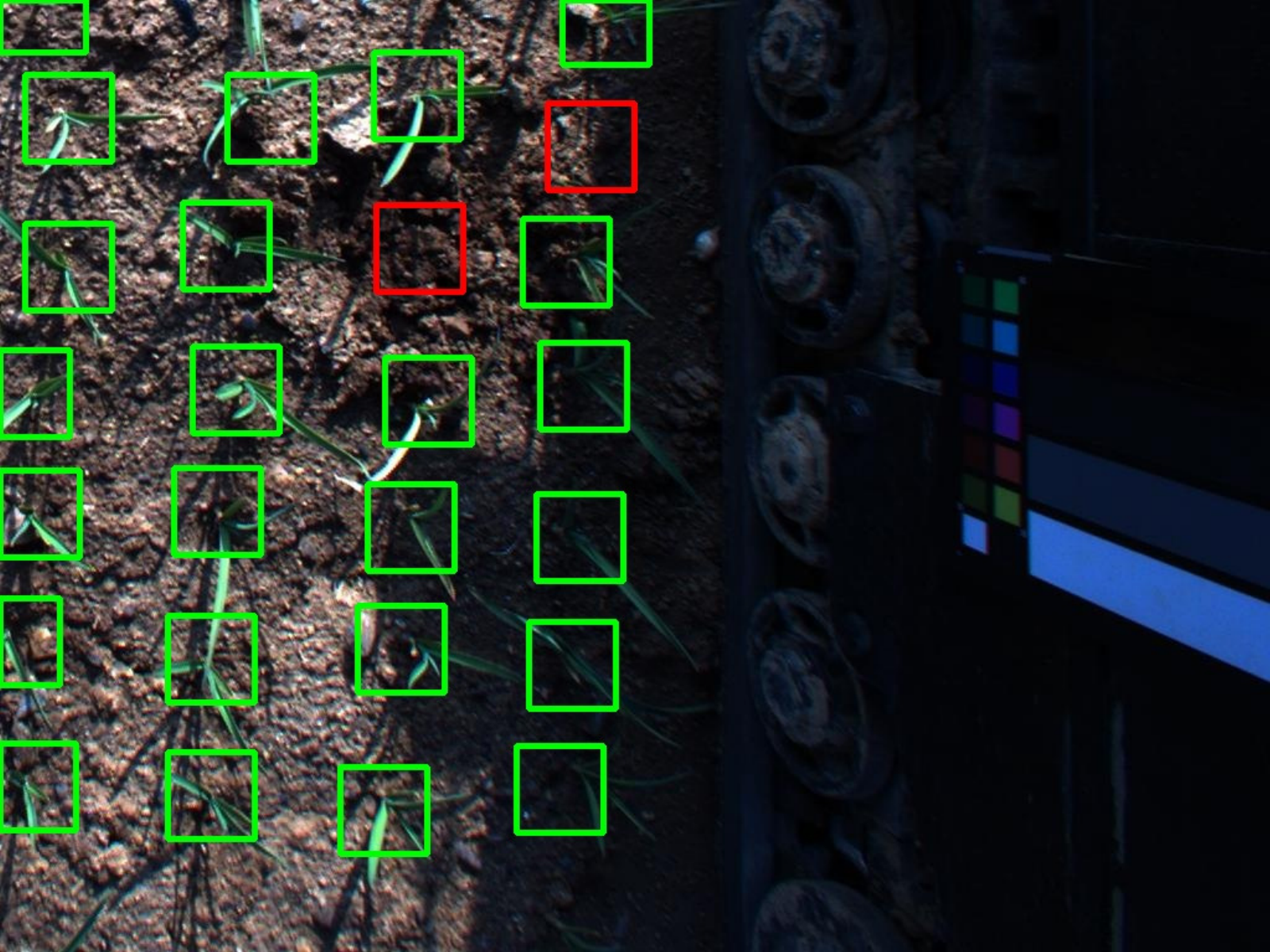}
\end{minipage}
\begin{minipage}{0.24\linewidth}
\centering
\includegraphics[width=\linewidth]{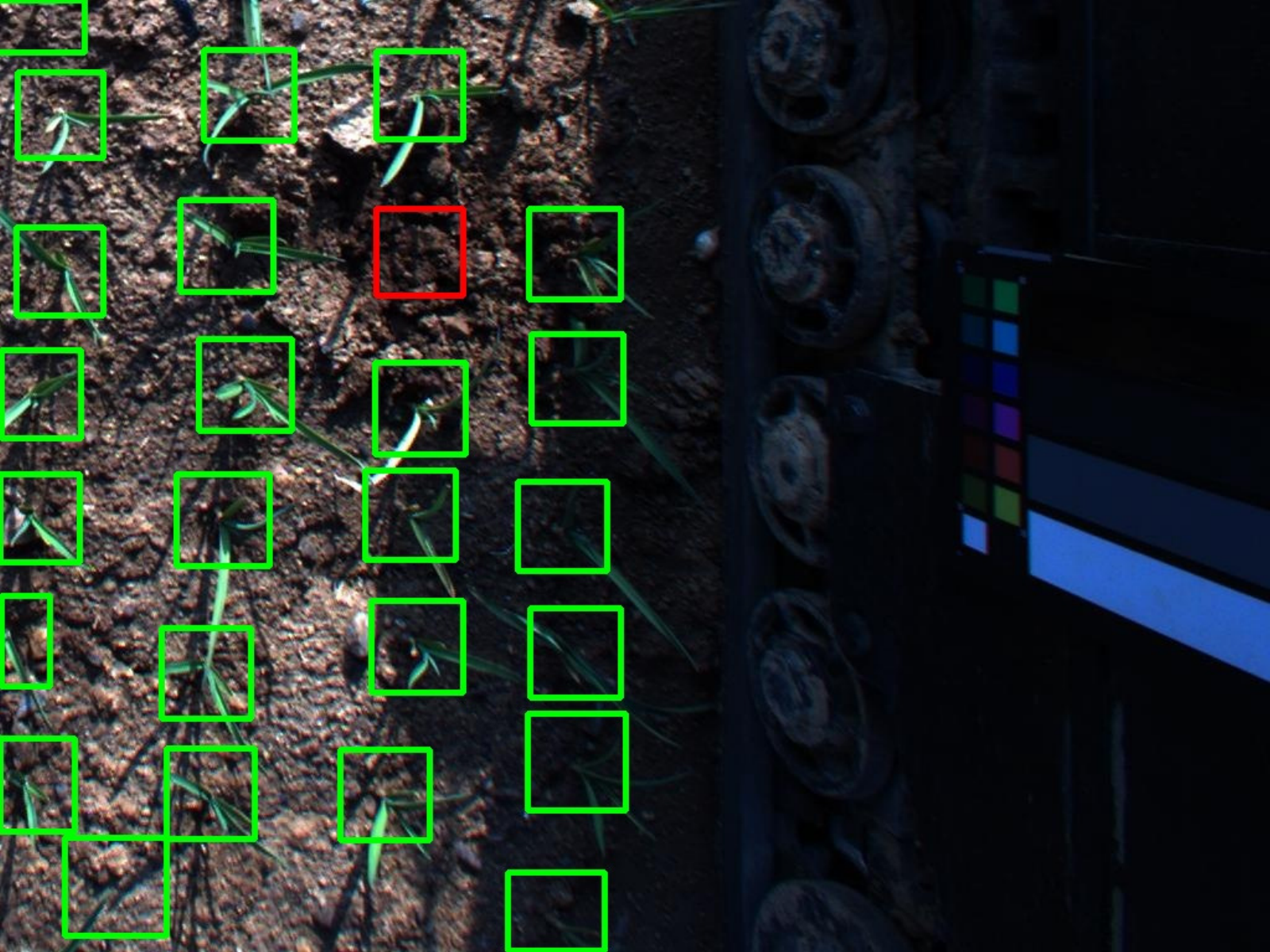}
\end{minipage}
\begin{minipage}{0.24\linewidth}
\centering
\includegraphics[width=\linewidth]{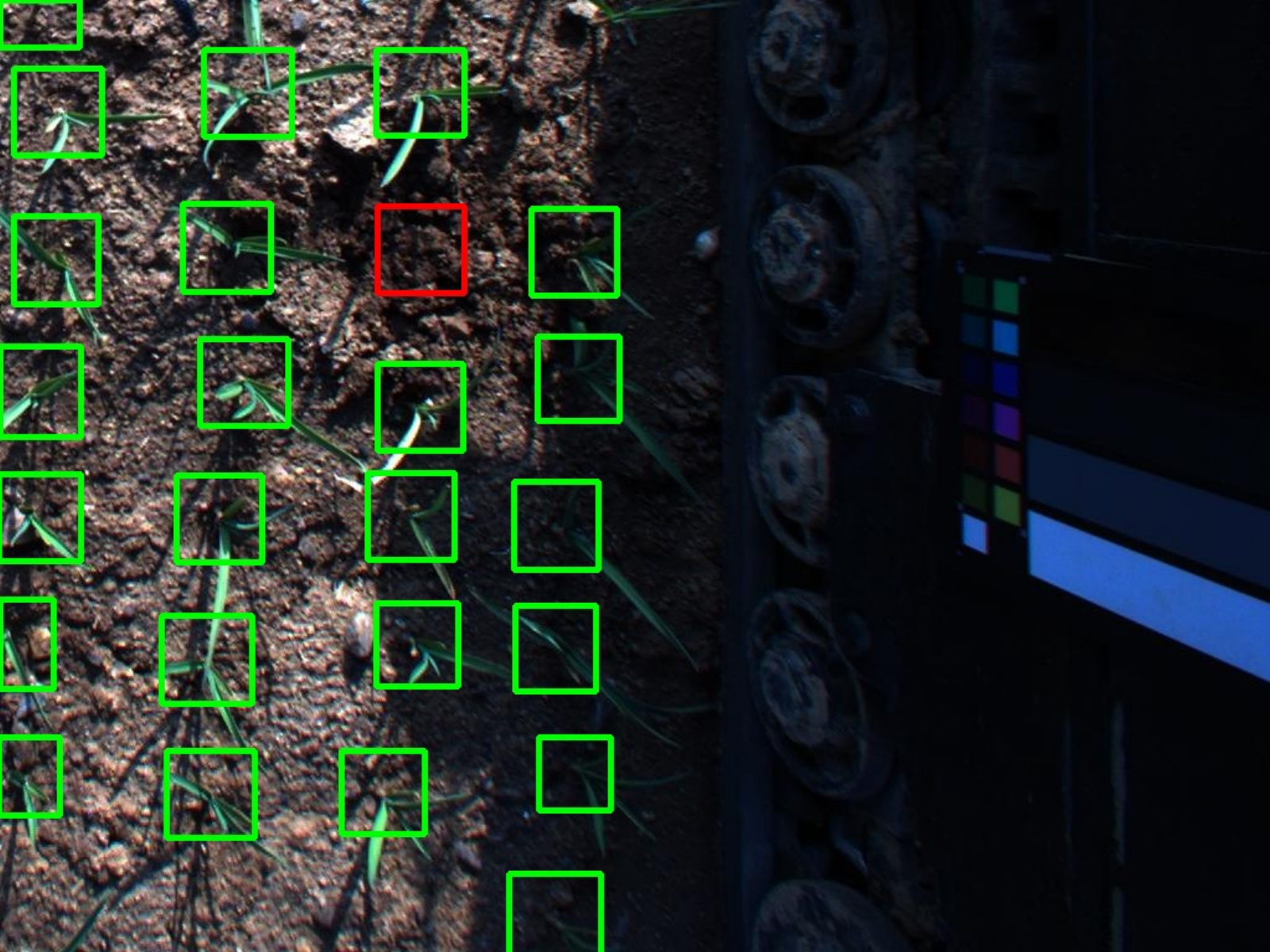}
\end{minipage}
\begin{minipage}{0.24\linewidth}
\centering
\includegraphics[width=\linewidth]{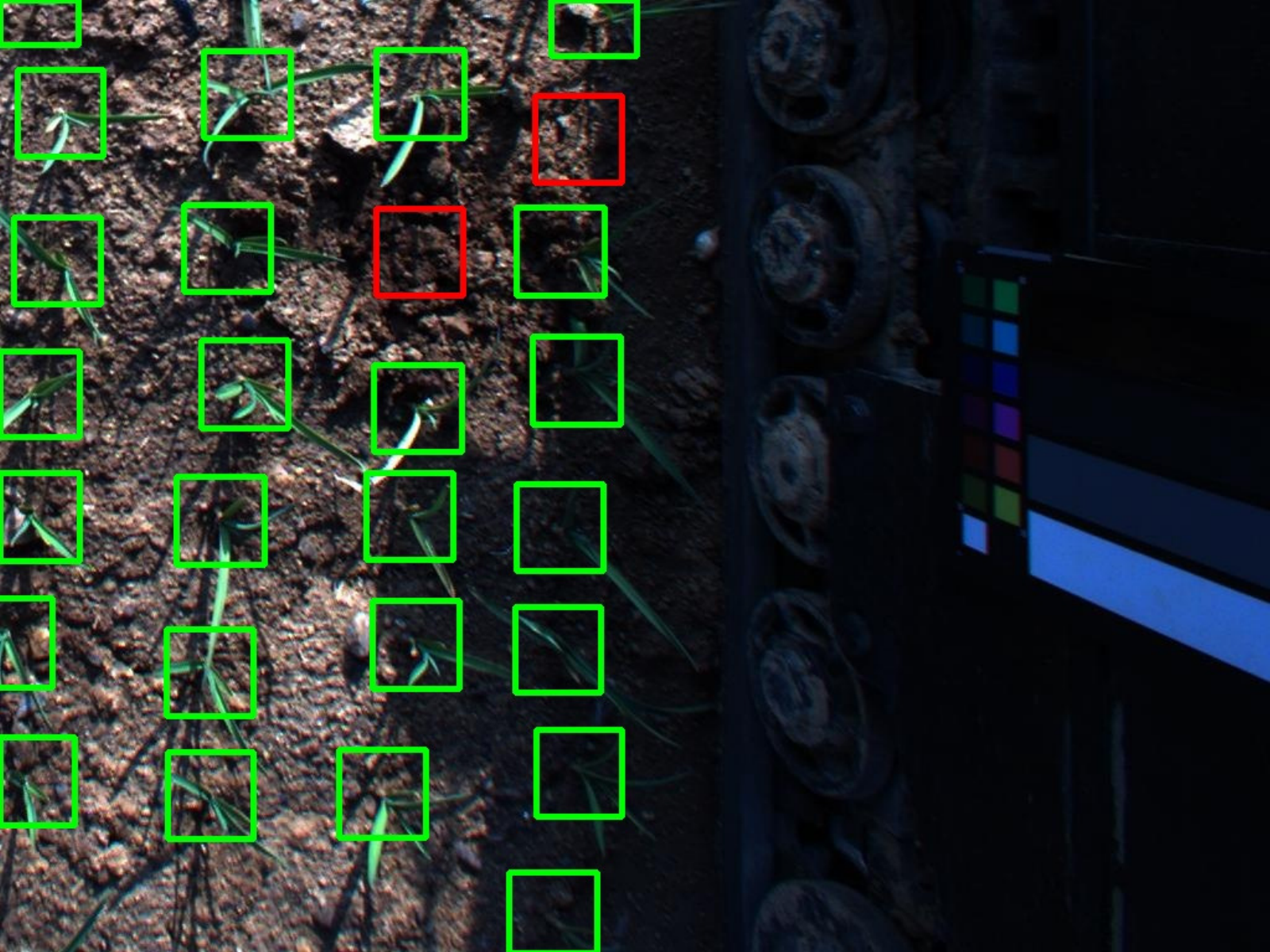}
\end{minipage}
\\
\vspace{0.05cm}

\begin{minipage}{0.24\linewidth}
\centering
\includegraphics[width=\linewidth]{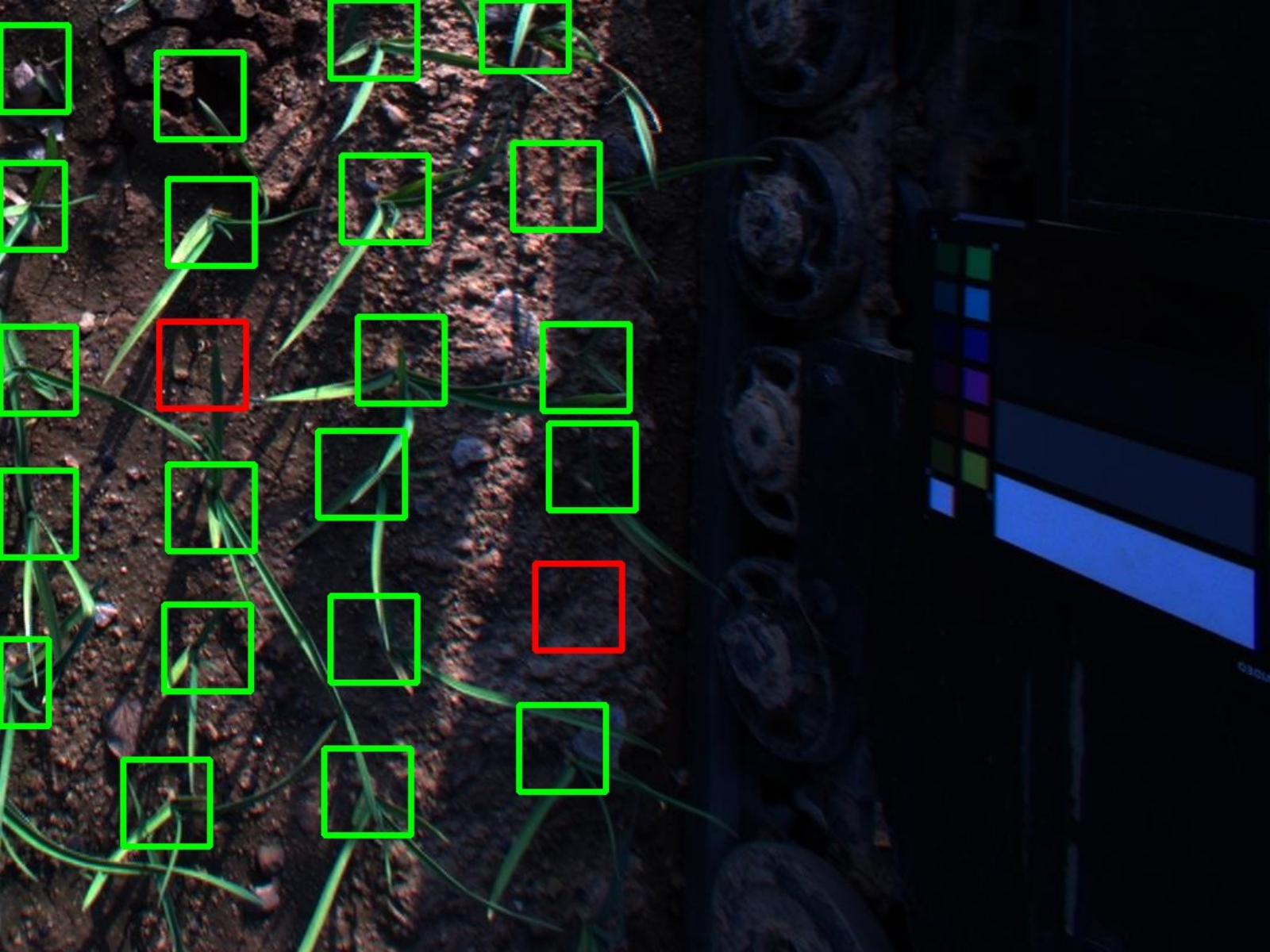}
\end{minipage}
\begin{minipage}{0.24\linewidth}
\centering
\includegraphics[width=\linewidth]{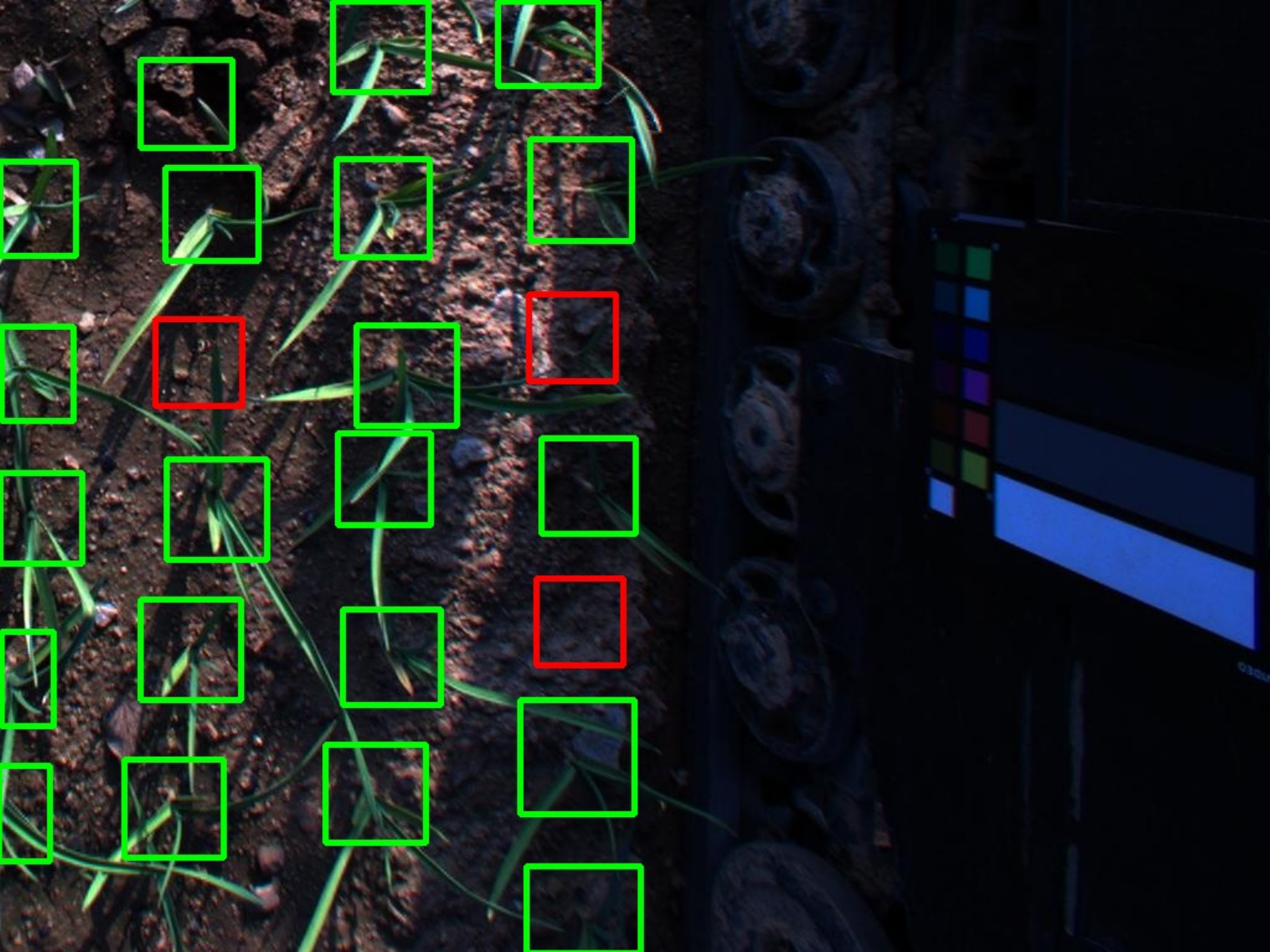}
\end{minipage}
\begin{minipage}{0.24\linewidth}
\centering
\includegraphics[width=\linewidth]{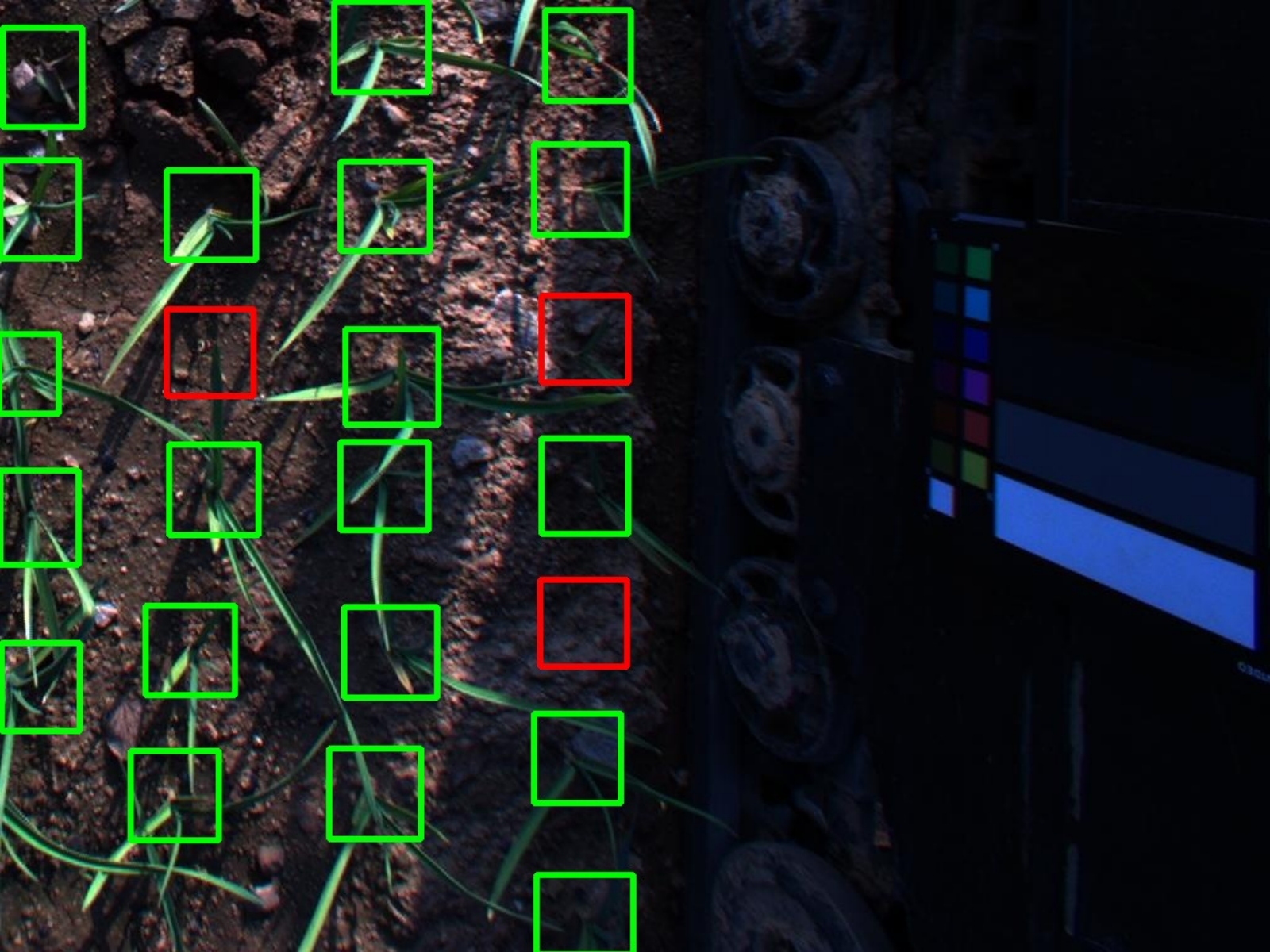}
\end{minipage}
\begin{minipage}{0.24\linewidth}
\centering
\includegraphics[width=\linewidth]{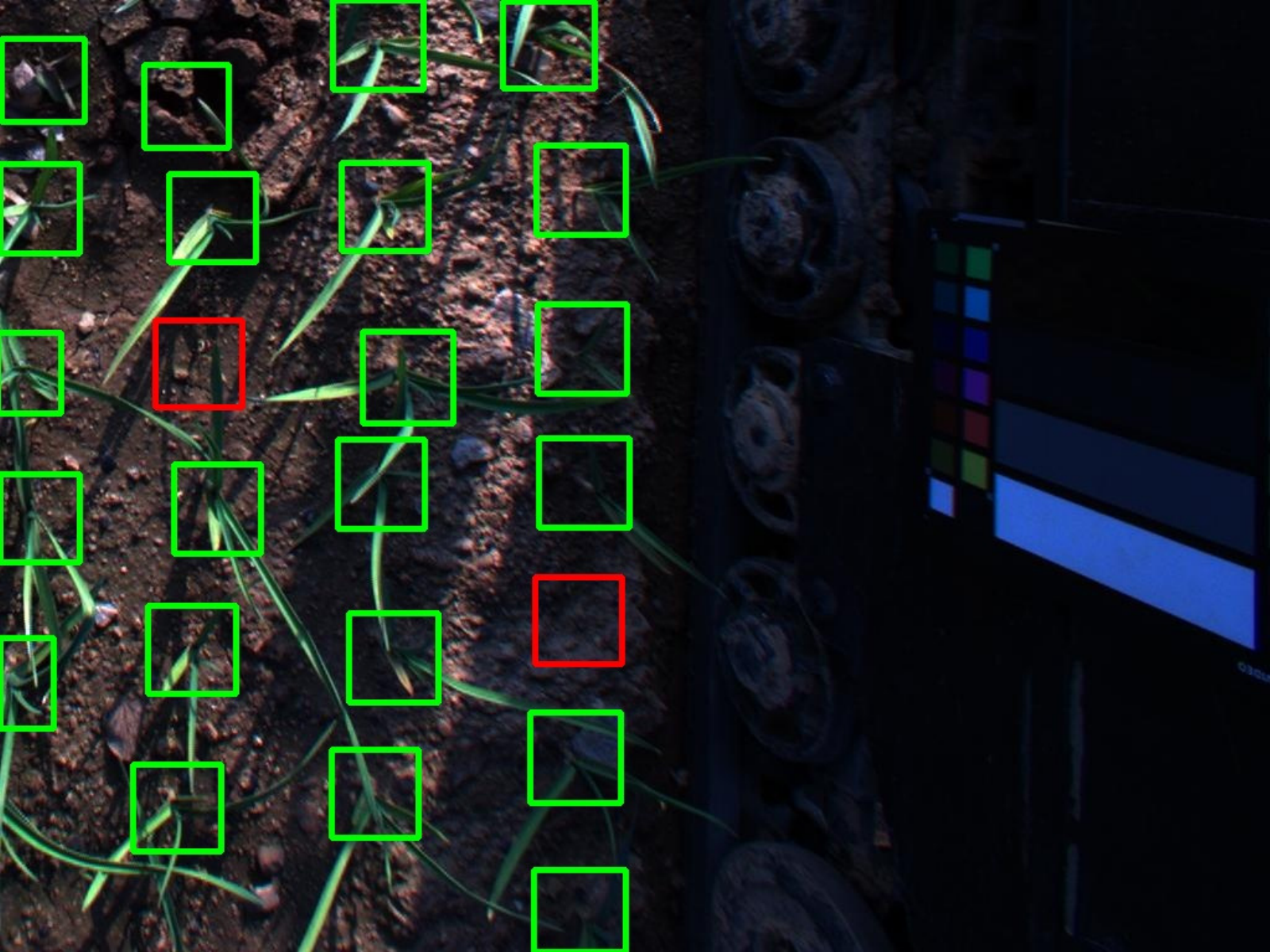}
\end{minipage}
\\
\vspace{0.05cm}

\begin{minipage}{0.24\linewidth}
\centering
\includegraphics[width=\linewidth]{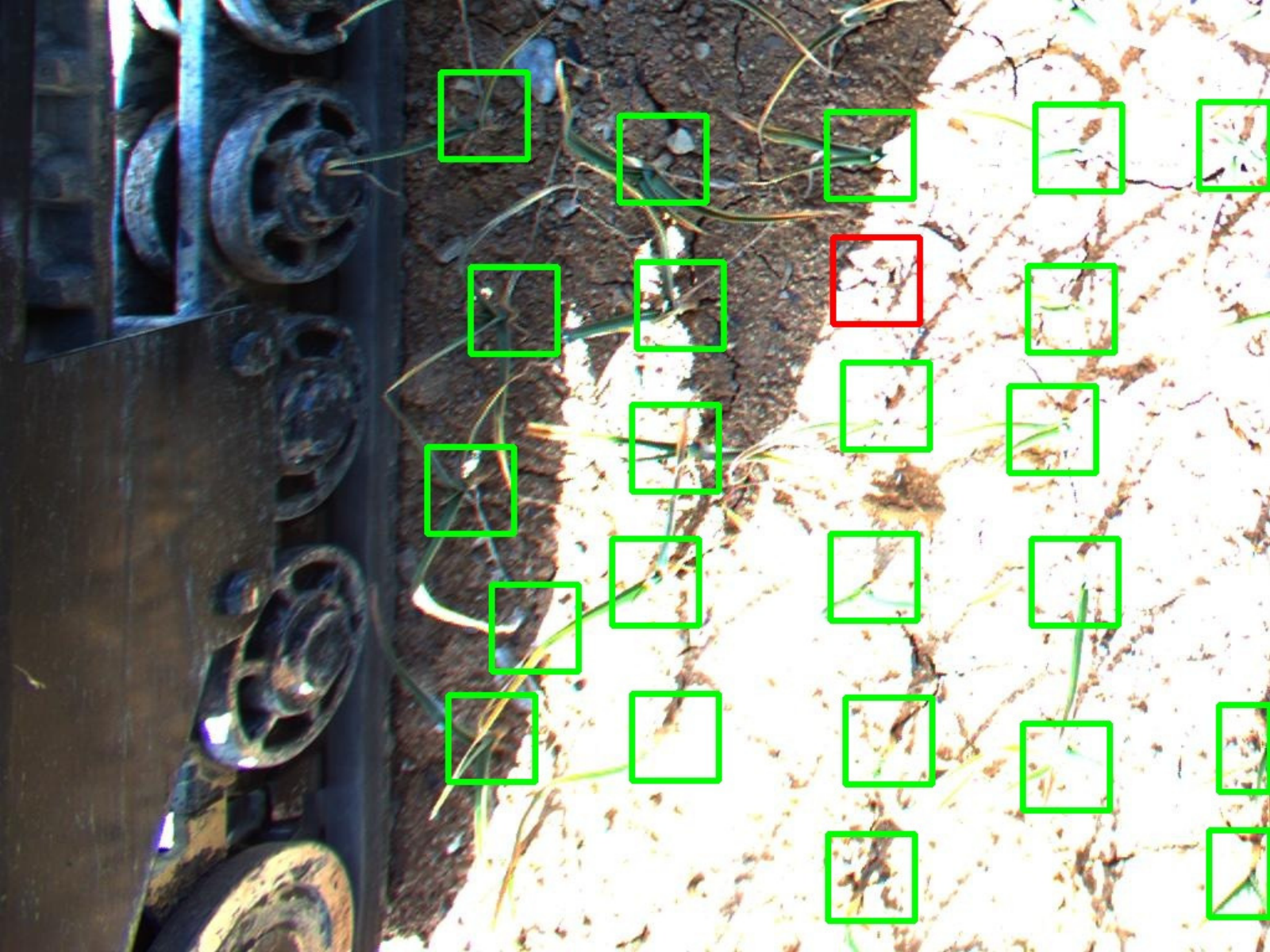}
\end{minipage}
\begin{minipage}{0.24\linewidth}
\centering
\includegraphics[width=\linewidth]{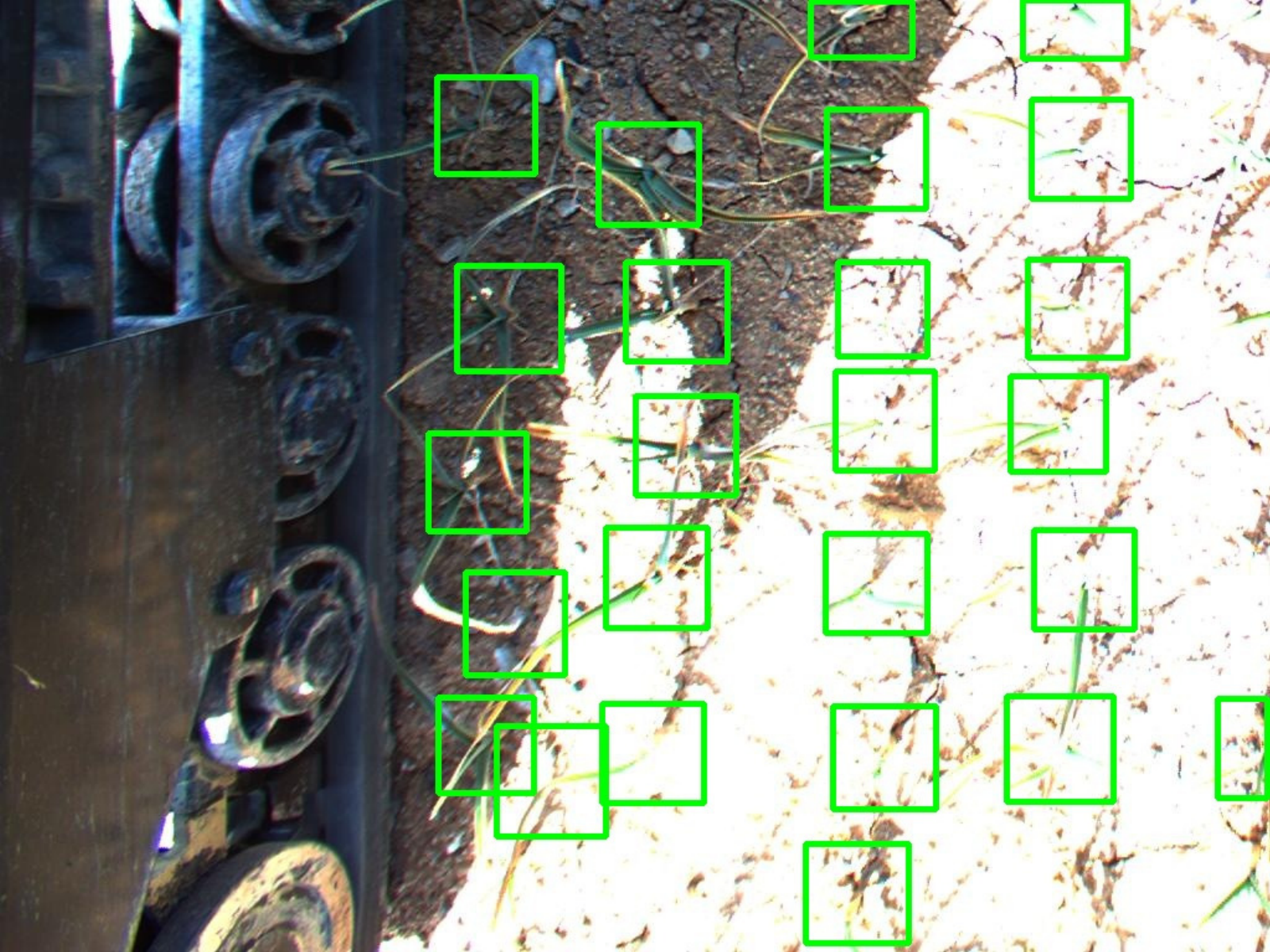}
\end{minipage}
\begin{minipage}{0.24\linewidth}
\centering
\includegraphics[width=\linewidth]{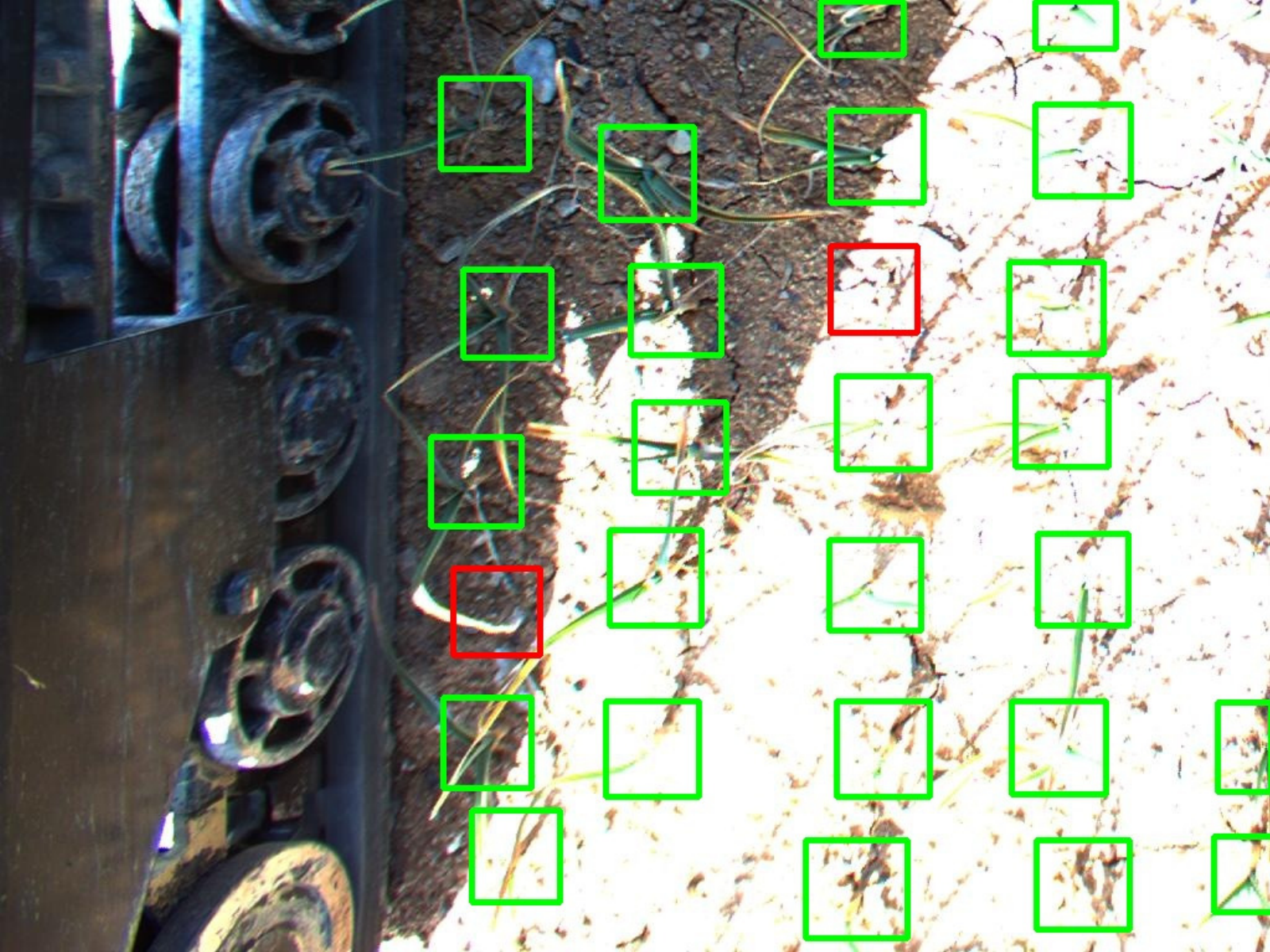}
\end{minipage}
\begin{minipage}{0.24\linewidth}
\centering
\includegraphics[width=\linewidth]{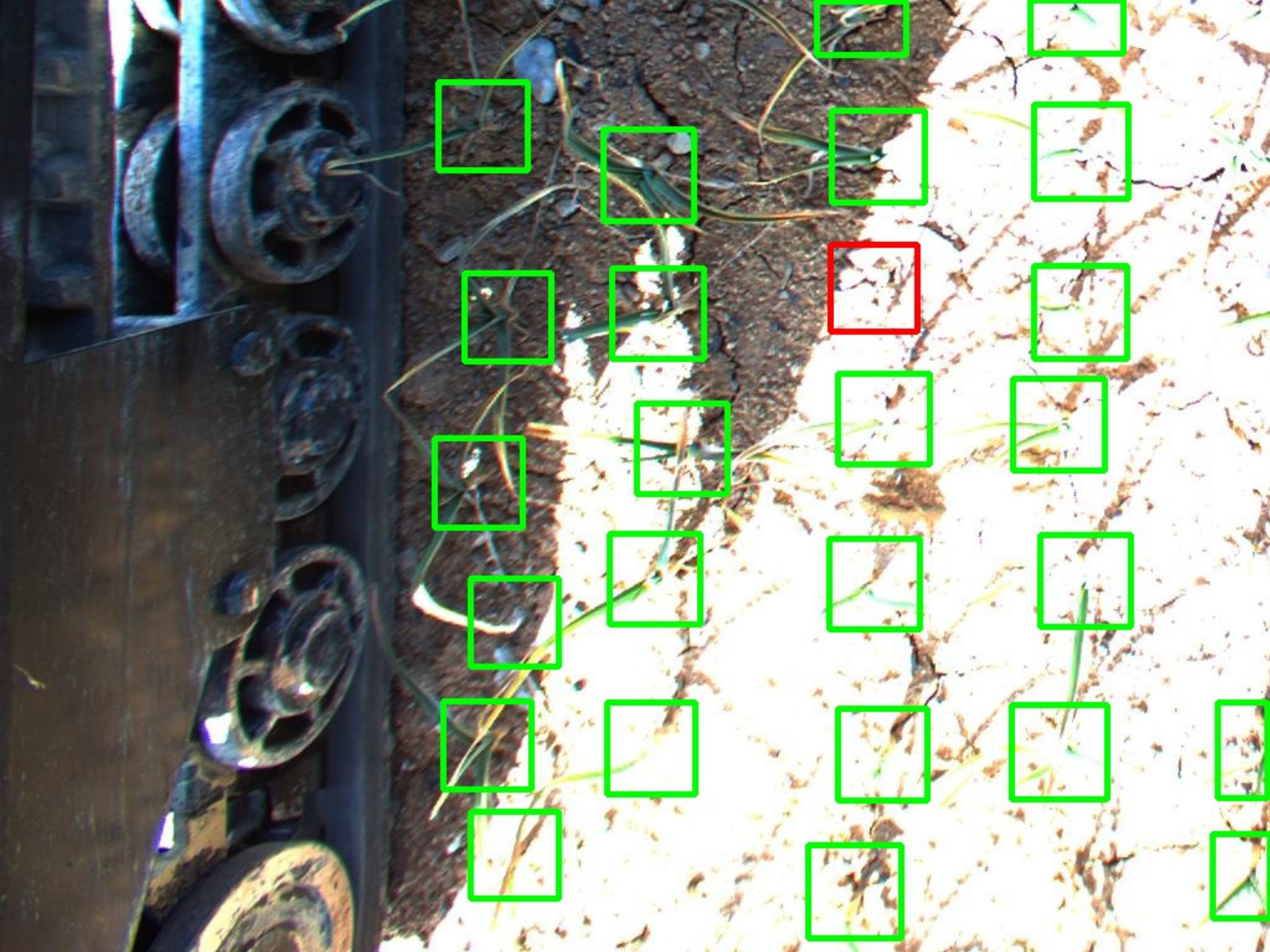}
\end{minipage}
\\
\vspace{0.05cm}

\begin{minipage}{0.24\linewidth}
\centering
(a) Ground Truth
\end{minipage}
\begin{minipage}{0.24\linewidth}
\centering
(b) DAI-Net
\end{minipage}
\begin{minipage}{0.24\linewidth}
\centering
(c) YOLA
\end{minipage}
\begin{minipage}{0.24\linewidth}
\centering
(d) Ours
\end{minipage}
\caption{
Qualitative comparison of seedling detection and missing seedling localization. (a) Ground truth, (b) results of DAI-Net~\citep{Du_2024_DAINet}, (c) results of YOLA~\citep{Hong_2024_YOLA}, and (d) results of the proposed method. Green and red boxes indicate detected locations of existing and missing seedlings, respectively.
}
\label{fig:result}
\end{figure*}

Figure~\ref{fig:result} presents qualitative comparisons of seedling detection and missing seedling localization between the proposed method and existing approaches, including DAI-Net~\citep{Du_2024_DAINet} and YOLA~\citep{Hong_2024_YOLA}. The proposed method produces more accurate and consistent seedling detections under challenging illumination conditions. In contrast, previous methods occasionally miss seedlings or generate inaccurate detections in regions affected by strong shadows or uneven lighting.

Such detection errors directly affect the subsequent missing seedling localization process. Missed detections may lead to incorrect identification of missing seedling locations, while false detections can disrupt the spatial distribution used for crop row fitting. As illustrated in Figure~\ref{fig:result}, the proposed method provides more reliable seedling predictions, which leads to more accurate estimation of missing seedling positions.

Overall, the qualitative results demonstrate that the proposed adversarial augmentation strategy improves the robustness of the detector and enables more stable missing seedling localization under real-world field conditions with complex illumination variations.

\begin{table}[t]
\centering
\caption{Ablation study on the augmentation policy agent and the design of its reward function.}
\label{tab:policy_learning_results}
\begin{minipage}{1\linewidth}
\centering
\begin{tabular}{>{\centering}m{0.2\textwidth} | >{\centering}m{0.2\textwidth} | >{\centering}m{0.2\textwidth} | >{\centering\arraybackslash}m{0.2\textwidth}}
\toprule
\multicolumn{3}{c|}{Method} & \multirow{2}{*}{AP$_{50}$} \\
\cmidrule{1-3}
$f_{\text{plc}}$ & $\mathcal{L}_{\text{det}}$ & $\mathcal{L}_{\text{struc}}$ & \\
\midrule
-- & -- & -- & 90.7 \\
\checkmark & -- & -- & 90.9 \\
\checkmark & \checkmark & -- & 91.0 \\
\checkmark & \checkmark & \checkmark & \textbf{91.6} \\
\bottomrule
\end{tabular}
\end{minipage}
\end{table}

\subsection{Analysis}
Unless otherwise stated, the analyses in this subsection are based on the YOLOv3 detector setting and the extreme-brightness stress-test split.

Table~\ref{tab:policy_learning_results} presents an ablation study on the effectiveness of the augmentation policy agent and the design of its reward function. The configuration without the augmentation policy agent corresponds to the ``Baseline''. Introducing the policy agent with a random reward improves the detection performance by 0.2\%.

When the reward is defined only by the detection loss $\mathcal{L}_{\text{det}}$, the performance further improves by 0.1\%. This limited gain indicates that simply encouraging the policy to generate samples with higher detection loss is not sufficient to achieve substantial robustness improvements. In this case, the policy may produce challenging samples, but they are not necessarily guaranteed to be structurally meaningful or representative of realistic illumination variations.

Finally, incorporating both the detection loss and the structural penalty $\mathcal{L}_{\text{struc}}$ into the reward yields an additional enhancement of 0.6\%, achieving the best AP$_{50}$ of 91.6\%. These results demonstrate that carefully designing the reward function is critical for effective augmentation learning. In particular, the structural penalty helps prevent excessive distortions and encourages the generation of challenging yet structurally consistent augmentations. This suggests that robustness is better promoted by balancing adversarial perturbation with structural preservation, rather than by simply maximizing the detection loss.

\begin{table}[t]
\centering
\caption{Effect of advantage normalization on REINFORCE-based policy learning.}\label{tab:adv_norm}
\begin{minipage}{1\linewidth}
\centering
\begin{tabular}{>{\centering}m{0.62\textwidth} | >{\centering\arraybackslash}m{0.25\textwidth}}
\toprule
Setting & AP$_{50}$ \\
\midrule
w/o advantage normalization & 91.4 \\
w/ advantage normalization & \textbf{91.6} \\
\bottomrule
\end{tabular}
\end{minipage}
\end{table}

We further examine the effect of advantage normalization on policy learning stability. As shown in Table~\ref{tab:adv_norm}, removing advantage normalization decreases AP$_{50}$ from 91.6 to 91.4. Although we did not observe severe instability such as divergence or policy collapse during training, the stochastic policy introduces reward fluctuations across rollouts. The advantage normalization in Eq.~(\ref{eq:advantage}) helps stabilize policy updates by normalizing reward values within the rollout set, while the periodic policy update interval $K$ reduces overly frequent policy changes. These results indicate that advantage normalization contributes to stable and effective REINFORCE-based policy optimization.

\begin{table}[t]
\centering
\caption{Effect of the hyperparameter $\beta_{\text{struc}}$ that balances $\mathcal{L}_{\text{det}}$ and $\mathcal{L}_{\text{struc}}$ in the reward function.}
\label{tab:beta_hard_results}
\begin{minipage}{1\linewidth}
\centering
\begin{tabular}{>{\centering}m{0.4\textwidth} | >{\centering\arraybackslash}m{0.4\textwidth}}
\toprule
$\beta_{\text{struc}}$ & AP$_{50}$ \\
\midrule
0 & 91.0 \\
5 & 91.3 \\
\textbf{10} & \textbf{91.6} \\
15 & 91.5 \\
20 & 91.5 \\
\bottomrule
\end{tabular}
\end{minipage}
\end{table}

\begin{table*}[t]
\centering
\caption{Inference-time cost-complexity and performance comparison with existing methods using YOLOv3. All measurements are conducted using the same input resolution on a single NVIDIA A5000 GPU.}
\label{tab:inference_complexity}
\begin{minipage}{1\linewidth}
\centering
\begin{tabular}{>{\centering}m{0.28\textwidth} | >{\centering}m{0.1\textwidth} >{\centering}m{0.1\textwidth} >{\centering}m{0.14\textwidth} | >{\centering}m{0.1\textwidth} >{\centering\arraybackslash}m{0.1\textwidth}}
\toprule
Method & Params (M) & GFLOPs & Inference time (ms/image) & AP$_{50}$ & F1-score \\
\midrule
Baseline & 61.524 & 55.18 & 19.8 & 90.7 & 65.0 \\
FeatEnHancer~\citep{Hashmi_2023_FeatEnHancer} & 61.663 & 86.73 & 20.5 & 91.3 & 64.4 \\
DAI-Net~\citep{Du_2024_DAINet} & 61.770 & 55.18 & 19.8 & 91.4 & 64.7 \\
YOLA~\citep{Hong_2024_YOLA} & 61.538 & 57.81 & 25.8 & 91.3 & 62.4 \\
\textbf{Ours} & 61.524 & 55.18 & 19.8 & \textbf{91.6} & \textbf{67.0} \\
\bottomrule
\end{tabular}
\end{minipage}
\end{table*}

\begin{table}[t]
\centering
\caption{Effect of different structural penalty terms in the reward function. L1 and SSIM penalties are evaluated using the same weight, $\beta_{\text{struc}}=10$.}
\label{tab:structural_penalty}
\begin{minipage}{1\linewidth}
\centering
\begin{tabular}{>{\centering}m{0.4\textwidth} | >{\centering\arraybackslash}m{0.4\textwidth}}
\toprule
Structural penalty & AP$_{50}$ \\
\midrule
None & 91.0 \\
L1 & 91.1 \\
SSIM & \textbf{91.6} \\
\bottomrule
\end{tabular}
\end{minipage}
\end{table}

Table~\ref{tab:beta_hard_results} analyzes the effect of the hyperparameter $\beta_{\text{struc}}$, which balances the detection loss $\mathcal{L}_{\text{det}}$ and the structural penalty $\mathcal{L}_{\text{struc}}$ in the reward function. Setting $\beta_{\text{struc}}=0$ removes the structural penalty, corresponding to the configuration reported in Table~\ref{tab:policy_learning_results}. As $\beta_{\text{struc}}$ increases from 0 to 10, the AP$_{50}$ improves from 91.0 to 91.6, indicating that moderate structural regularization helps the policy generate challenging yet structurally meaningful augmentations. When $\beta_{\text{struc}}$ is further increased to 15 and 20, the AP$_{50}$ slightly decreases and remains at 91.5 for both settings, which is still higher than the case without structural regularization. This result indicates that the proposed method benefits from structural regularization across the tested range, although overly strong structural constraints may slightly reduce augmentation diversity and weaken the adversarial effect. Therefore, $\beta_{\text{struc}}=10$ is selected as the best-performing value among the tested candidates, rather than as a theoretically guaranteed global optimum. Overall, these results confirm that balancing adversarial perturbation and structural preservation is important for effective policy learning.

We further compare SSIM with a pixel-wise L1 penalty to examine the choice of structural regularization. As shown in Table~\ref{tab:structural_penalty}, both L1 and SSIM penalties are evaluated using the same weight, $\beta_{\text{struc}}=10$. The L1-based penalty achieves an AP$_{50}$ of 91.1, which is slightly higher than the case without structural regularization but lower than the SSIM-based penalty. In contrast, the SSIM-based penalty achieves the best AP$_{50}$ of 91.6. This result suggests that enforcing strict pixel-wise similarity is less effective in our setting, where illumination-related variations should be allowed to improve robustness. SSIM provides a more suitable constraint by preserving perceptual and structural consistency while allowing moderate photometric changes.

We further analyze the operation selection frequency to examine how the policy uses the proposed augmentation space during training. Since the policy evolves during alternating optimization with the detector, we compute the selection frequency by accumulating the operations selected from the beginning of training up to the best-performing epoch. The most frequently selected operations are Full Image for region selection, Brightness for light adjustment, Motion Blur for texture detail, and Pixelate for drop out, with selection frequencies of 96.5\%, 87.5\%, 75.6\%, and 89.2\%, respectively. These results indicate that the policy learning process consistently emphasizes operations related to illumination changes and local visual degradation, which are closely associated with the challenging outdoor imaging conditions considered in this study. In particular, the frequent selection of Brightness supports the importance of illumination-related perturbations, while Motion Blur and Pixelate generate local appearance degradation that encourages the detector to learn more robust features. The frequent selection of Full Image suggests that applying perturbations broadly across the image is often effective in this dataset. Less frequently selected operations are not necessarily useless, because they may still be useful at specific training stages or for specific image conditions. Therefore, the large augmentation space provides diverse candidate transformations, while the policy learning process selectively exploits operations that are relevant to the input distribution and detection objective.

We analyze the inference-time cost-complexity of the proposed framework compared with existing methods. As shown in Table~\ref{tab:inference_complexity}, all measurements were conducted using the same input resolution on a single NVIDIA A5000 GPU. Since the policy network and augmentation module are used only during training and discarded during inference, the proposed method has the same inference-time complexity as the baseline detector, requiring 61.524M parameters, 55.18 GFLOPs, and 19.8 ms per image. Compared with existing illumination-robust methods, the proposed method achieves the best AP$_{50}$ of 91.6\% and the best missing seedling F1-score of 67.0\%, while maintaining lower or comparable inference-time complexity. FeatEnHancer~\citep{Hashmi_2023_FeatEnHancer} requires more computation with 86.73 GFLOPs and 20.5 ms per image, and YOLA~\citep{Hong_2024_YOLA} increases the inference time to 25.8 ms per image. Although DAI-Net~\citep{Du_2024_DAINet} has the same GFLOPs and inference time as the baseline in our implementation, its detection and downstream localization performance remain lower than those of the proposed method. These results indicate that the proposed framework improves robustness and downstream missing seedling localization reliability without increasing deployment-time model complexity.

\begin{table}[t]
\centering
\caption{Training-time cost comparison under the YOLOv3 detector setting. Training time is measured for 50 epochs on a single NVIDIA A5000 GPU.}
\label{tab:training_cost}
\begin{minipage}{1\linewidth}
\centering
\begin{tabular}{>{\centering}m{0.4\textwidth} | >{\centering\arraybackslash}m{0.4\textwidth}}
\toprule
Method & Training time (h) \\
\midrule
Baseline & 0.78 \\
Ours & 3.06 \\
\bottomrule
\end{tabular}
\end{minipage}
\end{table}

We also report the training-time cost of the proposed framework. As shown in Table~\ref{tab:training_cost}, the baseline detector required 0.78 h to train for 50 epochs, whereas the proposed framework required 3.06 h under the same setting, corresponding to an additional training time of 2.28 h. This increase is mainly caused by rollout-based policy optimization and alternating updates between the policy network and the detector. Nevertheless, the additional computation is incurred only during offline training, and the overall training time remains within a few hours in our experimental setting. During inference, only the detector is retained, so the additional policy learning cost does not affect deployment efficiency.

Although the proposed policy learning framework introduces additional training complexity through multiple policy heads, rollout-based reward estimation, and the structural penalty, these components are used only during training and do not introduce additional inference-time overhead. Therefore, the cost-benefit trade-off should be considered mainly in terms of training cost versus deployment reliability. While the AP$_{50}$ gain is modest, the proposed method improves missing seedling precision by 4.8 percentage points and F1-score by 2.0 percentage points compared with the baseline while maintaining the same recall, indicating more reliable downstream localization. Nevertheless, improving the efficiency of policy learning remains an important direction for future work.

\section{Conclusion and Limitation}
In this study, we proposed an illumination-robust seedling detection framework for ground-based agricultural monitoring under challenging outdoor lighting conditions. To improve detection robustness, we introduced an adversarial augmentation policy learning approach that jointly optimizes a stochastic policy agent and a seedling detector. The policy agent generates input-conditioned augmentations designed to challenge the detector during training. In addition, we incorporated a structural penalty into the reward function to prevent excessive distortions and preserve the structural consistency of the images.

To evaluate the proposed method, we constructed a new garlic seedling dataset captured using a ground-based monitoring platform under uncontrolled outdoor illumination conditions. Extensive experiments demonstrate that the proposed approach consistently improves seedling detection performance and leads to more reliable missing seedling localization compared with existing illumination-robust detection methods.

Despite its effectiveness, the proposed framework has several limitations. First, the policy learning process introduces additional computational overhead during training due to the use of multiple rollouts for reward estimation. Although this overhead does not affect inference-time efficiency, further simplifying the policy network and reducing the number of rollouts would improve the cost-benefit trade-off of the framework. Second, although the collected dataset contains a large number of annotated seedlings, the number of original field images remains relatively modest compared with large-scale object detection benchmarks. This may limit the scene-level diversity of the dataset in terms of field scenes, soil conditions, crop growth stages, and acquisition environments. In addition, we did not conduct cross-dataset evaluation because publicly available datasets with comparable ground-based outdoor seedling detection settings are not readily available. Future work will focus on improving the efficiency and scalability of policy learning and extending the approach to larger and more diverse agricultural datasets, as well as exploring generalization to other crop species and field conditions.

\section*{Acknowledgments}
This research was supported in part by the Korea Evaluation Institute of Industrial Technology (KEIT) Grant through the Korea Government (MOTIE) under Grant 20018635.

\bibliographystyle{elsarticle-harv}
\bibliography{mybibfile}

@INPROCEEDINGS{8014790,
  author={Tripathi, Subarna and Dane, Gokce and Kang, Byeongkeun and Bhaskaran, Vasudev and Nguyen, Truong},
  booktitle={2017 IEEE Conference on Computer Vision and Pattern Recognition Workshops (CVPRW)}, 
  title={LCDet: Low-Complexity Fully-Convolutional Neural Networks for Object Detection in Embedded Systems}, 
  year={2017},
  volume={},
  number={},
  pages={411-420},
  doi={10.1109/CVPRW.2017.56}}

@article{KANG2024108456,
title = {Improving weakly-supervised object localization using adversarial erasing and pseudo label},
journal = {Engineering Applications of Artificial Intelligence},
volume = {133},
pages = {108456},
year = {2024},
issn = {0952-1976},
doi = {https://doi.org/10.1016/j.engappai.2024.108456},
url = {https://www.sciencedirect.com/science/article/pii/S0952197624006146},
author = {Byeongkeun Kang and Sinhae Cha and Yeejin Lee},
}

@ARTICLE{Anandakrishnan_2024_LiYOLOv9,
  author={Anandakrishnan, Jayakrishnan and Sangaiah, Arun Kumar and Darmawan, Hendri and Son, Nguyen Khanh and Lin, Yi-Bing and Alenazi, Mohammed J. F.},
  journal={IEEE Journal of Selected Topics in Applied Earth Observations and Remote Sensing}, 
  title={Precise Spatial Prediction of Rice Seedlings From Large-Scale Airborne Remote Sensing Data Using Optimized Li-YOLOv9}, 
  year={2025},
  volume={18},
  number={},
  pages={2226-2238},
  doi={10.1109/JSTARS.2024.3505964}}

@article{PPO2017,
  author       = {John Schulman and
                  Filip Wolski and
                  Prafulla Dhariwal and
                  Alec Radford and
                  Oleg Klimov},
  title        = {Proximal Policy Optimization Algorithms},
  journal      = {CoRR},
  volume       = {abs/1707.06347},
  year         = {2017},
  url          = {http://arxiv.org/abs/1707.06347},
  eprinttype    = {arXiv},
  eprint       = {1707.06347},
  bibsource    = {dblp computer science bibliography, https://dblp.org}
}

@article{bakirci2025performance,
title = {Performance evaluation of low-power and lightweight object detectors for real-time monitoring in resource-constrained drone systems},
journal = {Engineering Applications of Artificial Intelligence},
volume = {159},
pages = {111775},
year = {2025},
issn = {0952-1976},
doi = {https://doi.org/10.1016/j.engappai.2025.111775},
url = {https://www.sciencedirect.com/science/article/pii/S0952197625017774},
author = {Murat Bakirci},
}

@ARTICLE{RAL_2022_Bang,
  author={Bang, Yeonjun and Lee, Yeejin and Kang, Byeongkeun},
  journal={IEEE Robotics and Automation Letters}, 
  title={Image-to-Image Translation-Based Data Augmentation for Robust EV Charging Inlet Detection}, 
  year={2022},
  volume={7},
  number={2},
  pages={3726-3733},
  doi={10.1109/LRA.2022.3146939}}

@InProceedings{Miao_2022_InstaAug,
  title = 	 {Learning Instance-Specific Augmentations by Capturing Local Invariances},
  author =       {Miao, Ning and Rainforth, Tom and Mathieu, Emile and Dubois, Yann and Teh, Yee Whye and Foster, Adam and Kim, Hyunjik},
  booktitle = 	 {Proceedings of the 40th International Conference on Machine Learning},
  pages = 	 {24720--24736},
  year = 	 {2023},
  editor = 	 {Krause, Andreas and Brunskill, Emma and Cho, Kyunghyun and Engelhardt, Barbara and Sabato, Sivan and Scarlett, Jonathan},
  volume = 	 {202},
  series = 	 {Proceedings of Machine Learning Research},
  month = 	 {23--29 Jul},
  publisher =    {PMLR},
  url = 	 {https://proceedings.mlr.press/v202/miao23a.html},
}

@software{yolov5,
title = {Ultralytics YOLOv5},
author = {Glenn Jocher},
year = {2020},
version = {7.0},
license = {AGPL-3.0},
url = {https://github.com/ultralytics/yolov5},
doi = {10.5281/zenodo.3908559},
orcid = {0000-0001-5950-6979}
}

@inproceedings{cheung_2022_adaaug,
title={AdaAug: Learning Class- and Instance-adaptive Data Augmentation Policies},
author={Tsz-Him Cheung and Dit-Yan Yeung},
booktitle={International Conference on Learning Representations},
year={2022},
url={https://openreview.net/forum?id=rWXfFogxRJN}
}

@INPROCEEDINGS{Minciullo_2021_DBGAN,
  author={Minciullo, Luca and Manhardt, Fabian and Yoshikawa, Kei and Meier, Sven and Tombari, Federico and Kobori, Norimasa},
  booktitle={2021 IEEE Winter Conference on Applications of Computer Vision (WACV)}, 
  title={DB-GAN: Boosting Object Recognition Under Strong Lighting Conditions}, 
  year={2021},
  volume={},
  number={},
  pages={2938-2948},
  doi={10.1109/WACV48630.2021.00298}}

@INPROCEEDINGS{Zhao2024DETRs,
  author={Zhao, Yian and Lv, Wenyu and Xu, Shangliang and Wei, Jinman and Wang, Guanzhong and Dang, Qingqing and Liu, Yi and Chen, Jie},
  booktitle={2024 IEEE/CVF Conference on Computer Vision and Pattern Recognition (CVPR)}, 
  title={DETRs Beat YOLOs on Real-time Object Detection}, 
  year={2024},
  volume={},
  number={},
  pages={16965-16974},
  doi={10.1109/CVPR52733.2024.01605}}

@misc{Tian2025YOLOv12,
      title={YOLOv12: Attention-Centric Real-Time Object Detectors}, 
      author={Yunjie Tian and Qixiang Ye and David Doermann},
      year={2025},
      eprint={2502.12524},
      archivePrefix={arXiv},
      primaryClass={cs.CV},
      url={https://arxiv.org/abs/2502.12524}, 
}

@InProceedings{Lin2014COCO,
author="Lin, Tsung-Yi
and Maire, Michael
and Belongie, Serge
and Hays, James
and Perona, Pietro
and Ramanan, Deva
and Doll{\'a}r, Piotr
and Zitnick, C. Lawrence",
editor="Fleet, David
and Pajdla, Tomas
and Schiele, Bernt
and Tuytelaars, Tinne",
title="Microsoft COCO: Common Objects in Context",
booktitle="Computer Vision -- ECCV 2014",
year="2014",
publisher="Springer International Publishing",
address="Cham",
pages="740--755",
isbn="978-3-319-10602-1"
}

@article{CHEN_2022_assimilation,
title = {An assimilation method for wheat failure detection at the seedling stage},
journal = {European Journal of Agronomy},
volume = {141},
pages = {126640},
year = {2022},
issn = {1161-0301},
doi = {https://doi.org/10.1016/j.eja.2022.126640},
url = {https://www.sciencedirect.com/science/article/pii/S1161030122001885},
author = {Pengfei Chen and Xiao Ma and Guijun Yang},
}

@Article{Gao_2022_design,
AUTHOR = {Gao, Junpeng and Li, Yuhua and Zhou, Kai and Wu, Yanqiang and Hou, Jialin},
TITLE = {Design and Optimization of a Machine-Vision-Based Complementary Seeding Device for Tray-Type Green Onion Seedling Machines},
JOURNAL = {Agronomy},
VOLUME = {12},
YEAR = {2022},
NUMBER = {9},
ARTICLE-NUMBER = {2180},
URL = {https://www.mdpi.com/2073-4395/12/9/2180},
ISSN = {2073-4395},
}

@Article{Gennaro_2020_evaluation,
AUTHOR = {Di Gennaro, Salvatore Filippo},
TITLE = {Evaluation of novel precision viticulture tool for canopy biomass estimation and missing plant detection based on 2.5D and 3D approaches using RGB images},
JOURNAL = {Plant Methods},
VOLUME = {16},
YEAR = {2020},
DOI = {https://doi.org/10.1186/s13007-020-00632-2},
SN = {1746-4811},
}

@article{Wu_2025_novel,
title = {A novel method for detecting missing seedlings based on UAV images and rice transplanter operation information},
journal = {Computers and Electronics in Agriculture},
volume = {229},
pages = {109789},
year = {2025},
issn = {0168-1699},
doi = {https://doi.org/10.1016/j.compag.2024.109789},
url = {https://www.sciencedirect.com/science/article/pii/S0168169924011803},
author = {Shuanglong Wu and Xingang Ma and Yuxuan Jin and Junda Yang and Wenhao Zhang and Hongming Zhang and Hailin Wang and Ying Chen and Caixia Lin and Long Qi},
}

@article{Cui_2023_real_time,
title = {Real-time missing seedling counting in paddy fields based on lightweight network and tracking-by-detection algorithm},
journal = {Computers and Electronics in Agriculture},
volume = {212},
pages = {108045},
year = {2023},
issn = {0168-1699},
doi = {https://doi.org/10.1016/j.compag.2023.108045},
url = {https://www.sciencedirect.com/science/article/pii/S0168169923004337},
author = {Jinrong Cui and Hong Zheng and Zhiwei Zeng and Yuling Yang and Ruijun Ma and Yuyuan Tian and Jianwei Tan and Xiao Feng and Long Qi},
}

@article{Yuan_2024_rapidly,
title = {Rapidly count crop seedling emergence based on waveform Method(WM) using drone imagery at the early stage},
journal = {Computers and Electronics in Agriculture},
volume = {220},
pages = {108867},
year = {2024},
issn = {0168-1699},
doi = {https://doi.org/10.1016/j.compag.2024.108867},
url = {https://www.sciencedirect.com/science/article/pii/S0168169924002588},
author = {Jie Yuan and Xu Li and Meng Zhou and Hengbiao Zheng and Zhitao Liu and Yang Liu and Ming Wen and Tao Cheng and Weixing Cao and Yan Zhu and Xia Yao},
}

@article{Li_2021_high_precision,
title = {A high-precision detection method of hydroponic lettuce seedlings status based on improved Faster RCNN},
journal = {Computers and Electronics in Agriculture},
volume = {182},
pages = {106054},
year = {2021},
issn = {0168-1699},
doi = {https://doi.org/10.1016/j.compag.2021.106054},
url = {https://www.sciencedirect.com/science/article/pii/S0168169921000727},
author = {Zhenbo Li and Ye Li and Yongbo Yang and Ruohao Guo and Jinqi Yang and Jun Yue and Yizhe Wang},
}

@article{YAN_2023_machine,
title = {Machine vision-based tomato plug tray missed seeding detection and empty cell replanting},
journal = {Computers and Electronics in Agriculture},
volume = {208},
pages = {107800},
year = {2023},
issn = {0168-1699},
doi = {https://doi.org/10.1016/j.compag.2023.107800},
url = {https://www.sciencedirect.com/science/article/pii/S0168169923001886},
author = {Zeyu Yan and Yiming Zhao and Weisong Luo and Xinting Ding and Kai Li and Zhi He and Yinggang Shi and Yongjie Cui}
}

@InProceedings{Zhang_2022_Bytetrack,
author="Zhang, Yifu
and Sun, Peize
and Jiang, Yi
and Yu, Dongdong
and Weng, Fucheng
and Yuan, Zehuan
and Luo, Ping
and Liu, Wenyu
and Wang, Xinggang",
editor="Avidan, Shai
and Brostow, Gabriel
and Ciss{\'e}, Moustapha
and Farinella, Giovanni Maria
and Hassner, Tal",
title="ByteTrack: Multi-object Tracking by Associating Every Detection Box",
booktitle="Computer Vision -- ECCV 2022",
year="2022",
publisher="Springer Nature Switzerland",
address="Cham",
pages="1--21",
isbn="978-3-031-20047-2"
}

@ARTICLE{Ren_2017_FasterRCNN,
  author={Ren, Shaoqing and He, Kaiming and Girshick, Ross and Sun, Jian},
  journal={IEEE Transactions on Pattern Analysis and Machine Intelligence}, 
  title={Faster R-CNN: Towards Real-Time Object Detection with Region Proposal Networks}, 
  year={2017},
  volume={39},
  number={6},
  pages={1137-1149},
  doi={10.1109/TPAMI.2016.2577031}}

@INPROCEEDINGS{Cui_2021_MAET,
  author={Cui, Ziteng and Qi, Guo-Jun and Gu, Lin and You, Shaodi and Zhang, Zenghui and Harada, Tatsuya},
  booktitle={2021 IEEE/CVF International Conference on Computer Vision (ICCV)}, 
  title={Multitask AET with Orthogonal Tangent Regularity for Dark Object Detection}, 
  year={2021},
  volume={},
  number={},
  pages={2533-2542},
  doi={10.1109/ICCV48922.2021.00255}}

@article{Liu_2022_IAYolo, 
title={Image-Adaptive YOLO for Object Detection in Adverse Weather Conditions}, 
volume={36}, 
url={https://ojs.aaai.org/index.php/AAAI/article/view/20072}, 
DOI={10.1609/aaai.v36i2.20072}, 
number={2}, 
journal={Proceedings of the AAAI Conference on Artificial Intelligence}, 
author={Liu, Wenyu and Ren, Gaofeng and Yu, Runsheng and Guo, Shi and Zhu, Jianke and Zhang, Lei}, 
year={2022}, 
month={Jun.},
pages={1792-1800} 
}

@INPROCEEDINGS{Kalwar_2023_GDIP,
  author={Kalwar, Sanket and Patel, Dhruv and Aanegola, Aakash and Konda, Krishna Reddy and Garg, Sourav and Krishna, K Madhava},
  booktitle={2023 IEEE International Conference on Robotics and Automation (ICRA)}, 
  title={GDIP: Gated Differentiable Image Processing for Object Detection in Adverse Conditions}, 
  year={2023},
  volume={},
  number={},
  pages={7083-7089},
  doi={10.1109/ICRA48891.2023.10160356}
}

@ARTICLE{XI_2024_DroneDet,
  author={Xi, Yue and Jia, Wenjing and Miao, Qiguang and Feng, Junmei and Ren, Jinchang and Luo, Heng},
  journal={IEEE Transactions on Geoscience and Remote Sensing}, 
  title={Detection-Driven Exposure-Correction Network for Nighttime Drone-View Object Detection}, 
  year={2024},
  volume={62},
  number={},
  pages={1-14},
  doi={10.1109/TGRS.2024.3351134}}

@INPROCEEDINGS{Hashmi_2023_FeatEnHancer,
  author={Hashmi, Khurram Azeem and Kallempudi, Goutham and Stricker, Didier and Afzal, Muhammamd Zeshan},
  booktitle={2023 IEEE/CVF International Conference on Computer Vision (ICCV)}, 
  title={FeatEnHancer: Enhancing Hierarchical Features for Object Detection and Beyond Under Low-Light Vision}, 
  year={2023},
  volume={},
  number={},
  pages={6702-6712},
  doi={10.1109/ICCV51070.2023.00619}}

@inproceedings{Hong_2024_YOLA,
 author = {Hong, Mingbo and Cheng, Shen and Huang, Haibin and Fan, Haoqiang and Liu, Shuaicheng},
 booktitle = {Advances in Neural Information Processing Systems},
 doi = {10.52202/079017-2765},
 editor = {A. Globerson and L. Mackey and D. Belgrave and A. Fan and U. Paquet and J. Tomczak and C. Zhang},
 pages = {87136--87158},
 publisher = {Curran Associates, Inc.},
 title = {You Only Look Around: Learning Illumination-Invariant Feature for Low-light Object Detection},
 url = {https://proceedings.neurips.cc/paper_files/paper/2024/file/9e74900c3f6100c56add4bf417547848-Paper-Conference.pdf},
 volume = {37},
 year = {2024}
}

@INPROCEEDINGS{Du_2024_DAINet,
  author={Du, Zhipeng and Shit, Miaojing and Deng, Jiankang},
  booktitle={2024 IEEE/CVF Conference on Computer Vision and Pattern Recognition (CVPR)}, 
  title={Boosting Object Detection with Zero-Shot Day-Night Domain Adaptation}, 
  year={2024},
  volume={},
  number={},
  pages={12666-12676},
  doi={10.1109/CVPR52733.2024.01204}}

@INPROCEEDINGS{Guo_2020_ZeroDCE,
  author={Guo, Chunle and Li, Chongyi and Guo, Jichang and Loy, Chen Change and Hou, Junhui and Kwong, Sam and Cong, Runmin},
  booktitle={2020 IEEE/CVF Conference on Computer Vision and Pattern Recognition (CVPR)}, 
  title={Zero-Reference Deep Curve Estimation for Low-Light Image Enhancement}, 
  year={2020},
  volume={},
  number={},
  pages={1777-1786},
  doi={10.1109/CVPR42600.2020.00185}
  }

@INPROCEEDINGS{Liu_2021_RUAS,
  author={Liu, Risheng and Ma, Long and Zhang, Jiaao and Fan, Xin and Luo, Zhongxuan},
  booktitle={2021 IEEE/CVF Conference on Computer Vision and Pattern Recognition (CVPR)}, 
  title={Retinex-inspired Unrolling with Cooperative Prior Architecture Search for Low-light Image Enhancement}, 
  year={2021},
  volume={},
  number={},
  pages={10556-10565},
  doi={10.1109/CVPR46437.2021.01042}}

@INPROCEEDINGS{Ma_2022_SCI,
  author={Ma, Long and Ma, Tengyu and Liu, Risheng and Fan, Xin and Luo, Zhongxuan},
  booktitle={2022 IEEE/CVF Conference on Computer Vision and Pattern Recognition (CVPR)}, 
  title={Toward Fast, Flexible, and Robust Low-Light Image Enhancement}, 
  year={2022},
  volume={},
  number={},
  pages={5627-5636},
  doi={10.1109/CVPR52688.2022.00555}}

@InProceedings{Chen_2018_RetinexNet,
 title={Deep Retinex Decomposition for Low-Light Enhancement},
 author={Wei, Chen and Wang, Wenjing and Yang, Wenhan and Liu, Jiaying}, 
 booktitle={British Machine Vision Conference},
 year={2018},
 organization={British Machine Vision Association}
}

@inbook{Shafer_1992_Lambertian,
author = {Shafer, Steven A.},
title = {Using color to separate reflection components},
year = {1992},
isbn = {0867202955},
publisher = {Jones and Bartlett Publishers, Inc.},
address = {USA},
booktitle = {Color},
pages = {43–51},
numpages = {9}
}

@INPROCEEDINGS{Cubuk_2019_AutoAugment,
  author={Cubuk, Ekin D. and Zoph, Barret and Mané, Dandelion and Vasudevan, Vijay and Le, Quoc V.},
  booktitle={2019 IEEE/CVF Conference on Computer Vision and Pattern Recognition (CVPR)}, 
  title={AutoAugment: Learning Augmentation Strategies From Data}, 
  year={2019},
  volume={},
  number={},
  pages={113-123},
  doi={10.1109/CVPR.2019.00020}}

@INPROCEEDINGS{Lin_2019_OHL_AutoAug,
  author={Lin, Chen and Guo, Minghao and Li, Chuming and Yuan, Xin and Wu, Wei and Yan, Junjie and Lin, Dahua and Ouyang, Wanli},
  booktitle={2019 IEEE/CVF International Conference on Computer Vision (ICCV)}, 
  title={Online Hyper-Parameter Learning for Auto-Augmentation Strategy}, 
  year={2019},
  volume={},
  number={},
  pages={6578-6587},
  doi={10.1109/ICCV.2019.00668}}

@Article{Li_2025_AROID,
author={Li, Lin
and Qiu, Jianing
and Spratling, Michael},
title={AROID: Improving Adversarial Robustness Through Online Instance-Wise Data Augmentation},
journal={International Journal of Computer Vision},
year={2025},
month={Feb},
day={01},
volume={133},
number={2},
pages={929-950},
issn={1573-1405},
doi={10.1007/s11263-024-02206-4},
url={https://doi.org/10.1007/s11263-024-02206-4}
}

@inproceedings{Zhang_2020_Adversarial_Autoaugm,
title={Adversarial AutoAugment},
author={Xinyu Zhang and Qiang Wang and Jian Zhang and Zhao Zhong},
booktitle={International Conference on Learning Representations},
year={2020},
url={https://openreview.net/forum?id=ByxdUySKvS}
}

@Article{Williams_REINFORCE_1992b,
author={Williams, Ronald J.},
title={Simple statistical gradient-following algorithms for connectionist reinforcement learning},
journal={Machine Learning},
year={1992},
month={May},
day={01},
volume={8},
number={3},
pages={229-256},
issn={1573-0565},
doi={10.1007/BF00992696},
url={https://doi.org/10.1007/BF00992696}
}

@InProceedings{He_PRN18_2016,
author="He, Kaiming
and Zhang, Xiangyu
and Ren, Shaoqing
and Sun, Jian",
editor="Leibe, Bastian
and Matas, Jiri
and Sebe, Nicu
and Welling, Max",
title="Identity Mappings in Deep Residual Networks",
booktitle="Computer Vision -- ECCV 2016",
year="2016",
publisher="Springer International Publishing",
address="Cham",
pages="630--645",
isbn="978-3-319-46493-0"
}

@INPROCEEDINGS{Feng_2021_tood,
  author={Feng, Chengjian and Zhong, Yujie and Gao, Yu and Scott, Matthew R. and Huang, Weilin},
  booktitle={2021 IEEE/CVF International Conference on Computer Vision (ICCV)}, 
  title={TOOD: Task-aligned One-stage Object Detection}, 
  year={2021},
  volume={},
  number={},
  pages={3490-3499},
  doi={10.1109/ICCV48922.2021.00349}}

@article{Redmon_2018_yolov3,
  author       = {Joseph Redmon and
                  Ali Farhadi},
  title        = {YOLOv3: An Incremental Improvement},
  journal      = {CoRR},
  volume       = {abs/1804.02767},
  year         = {2018},
  url          = {http://arxiv.org/abs/1804.02767},
  eprinttype    = {arXiv},
  eprint       = {1804.02767},
  bibsource    = {dblp computer science bibliography, https://dblp.org}
}

@INPROCEEDINGS{Guo_2010_exposure_map,
  author={Guo, Dong and Cheng, Yuan and Zhuo, Shaojie and Sim, Terence},
  booktitle={2010 IEEE Computer Society Conference on Computer Vision and Pattern Recognition}, 
  title={Correcting over-exposure in photographs}, 
  year={2010},
  volume={},
  number={},
  pages={515-521},
  doi={10.1109/CVPR.2010.5540170}}

\end{document}